\documentclass{article}
\usepackage{colm2024_conference}
\usepackage{listings}
\usepackage{booktabs}
\usepackage{graphicx}
\usepackage{enumitem}
\usepackage{wrapfig}
\usepackage[breakable]{tcolorbox}
\usepackage{algorithm}
\usepackage{algpseudocode}
\usepackage{verbatim} 
\usepackage{siunitx}
\usepackage{graphicx} 
\usepackage{tabularx} 
\usepackage{subcaption} 
\usepackage{multirow}    
\usepackage{arydshln}
\usepackage{xcolor}
\usepackage{wrapfig}

\usepackage{amsmath,amsfonts,bm}

\def\eqref#1{equation~\ref{#1}}

\def\1{\bm{1}}

\DeclareMathAlphabet{\mathsfit}{\encodingdefault}{\sfdefault}{m}{sl}
\SetMathAlphabet{\mathsfit}{bold}{\encodingdefault}{\sfdefault}{bx}{n}

\title{WorldPM: Scaling Human Preference Modeling}

\author{ \small
    Binghai Wang\textsuperscript{1,2$\ddagger$}, 
    Runji Lin\textsuperscript{1$\dagger$}, 
    Keming Lu\textsuperscript{1$\dagger$}, 
    Le Yu\textsuperscript{1}, 
    Zhenru Zhang\textsuperscript{1}, 
    Fei Huang\textsuperscript{1$\ast$}, 
    Chujie Zheng\textsuperscript{1}, 
    Kai Dang\textsuperscript{1}, 
    Yang Fan\textsuperscript{1}, 
    Xingzhang Ren\textsuperscript{1}, 
    An Yang\textsuperscript{1}, 
    Binyuan Hui\textsuperscript{1}, 
    Dayiheng Liu\textsuperscript{1},
    Tao Gui\textsuperscript{2$\ast$}, 
    Qi Zhang\textsuperscript{2}, 
    Xuanjing Huang\textsuperscript{2}, 
    Yu-Gang Jiang\textsuperscript{2},
    Bowen Yu\textsuperscript{1$\ast$}, 
    Jingren Zhou\textsuperscript{1},
    Junyang Lin\textsuperscript{1}
    \\
    \vspace{0.5em}
    \textbf{\textsuperscript{1}Qwen Team, Alibaba Group} \quad \textbf{\textsuperscript{2}Fudan University} 
    \vspace{-2.5em}
}

\footnotetext[0]{$^\ddagger$ Work done during the internship at Qwen. $^\ast$Corresponding authors: tgui@fudan.edu.cn, feihu.hf@alibaba-inc.com, yubowen.ybw@alibaba-inc.com. $^\dagger$Work done while working at Qwen.}

\begin{document}

\maketitle

\begin{abstract}
Motivated by scaling laws in language modeling that demonstrate how test loss scales as a power law with model and dataset sizes, we find that similar laws exist in preference modeling.
We propose \textbf{World} \textbf{P}reference \textbf{M}odeling (WorldPM) to emphasize this scaling potential, where World Preference embodies a unified representation of human preferences.
In this paper, we collect preference data from public forums covering diverse user communities, and conduct extensive training using 15M-scale data across models ranging from 1.5B to 72B parameters. 
We observe distinct patterns across different evaluation metrics: (1) Adversarial metrics (ability to identify deceptive features) consistently scale up with increased training data and base model size; (2) Objective metrics (objective knowledge with well-defined answers) show emergent behavior in larger language models, highlighting WorldPM's scalability potential; (3) Subjective metrics (subjective preferences from a limited number of humans or AI) do not demonstrate scaling trends.
Further experiments validate the effectiveness of WorldPM as a foundation for preference fine-tuning. Through evaluations on 7 benchmarks with 20 subtasks, we find that WorldPM broadly improves the generalization performance across human preference datasets of varying sizes (7K, 100K and 800K samples), with performance gains exceeding 5\% on many key subtasks. Integrating WorldPM into our internal RLHF pipeline, we observe significant improvements on both in-house and public evaluation sets, with notable gains of 4\% to 8\% in our in-house evaluations.

\begin{figure}[!hb]
    \centering
    \includegraphics[width=0.95\linewidth]{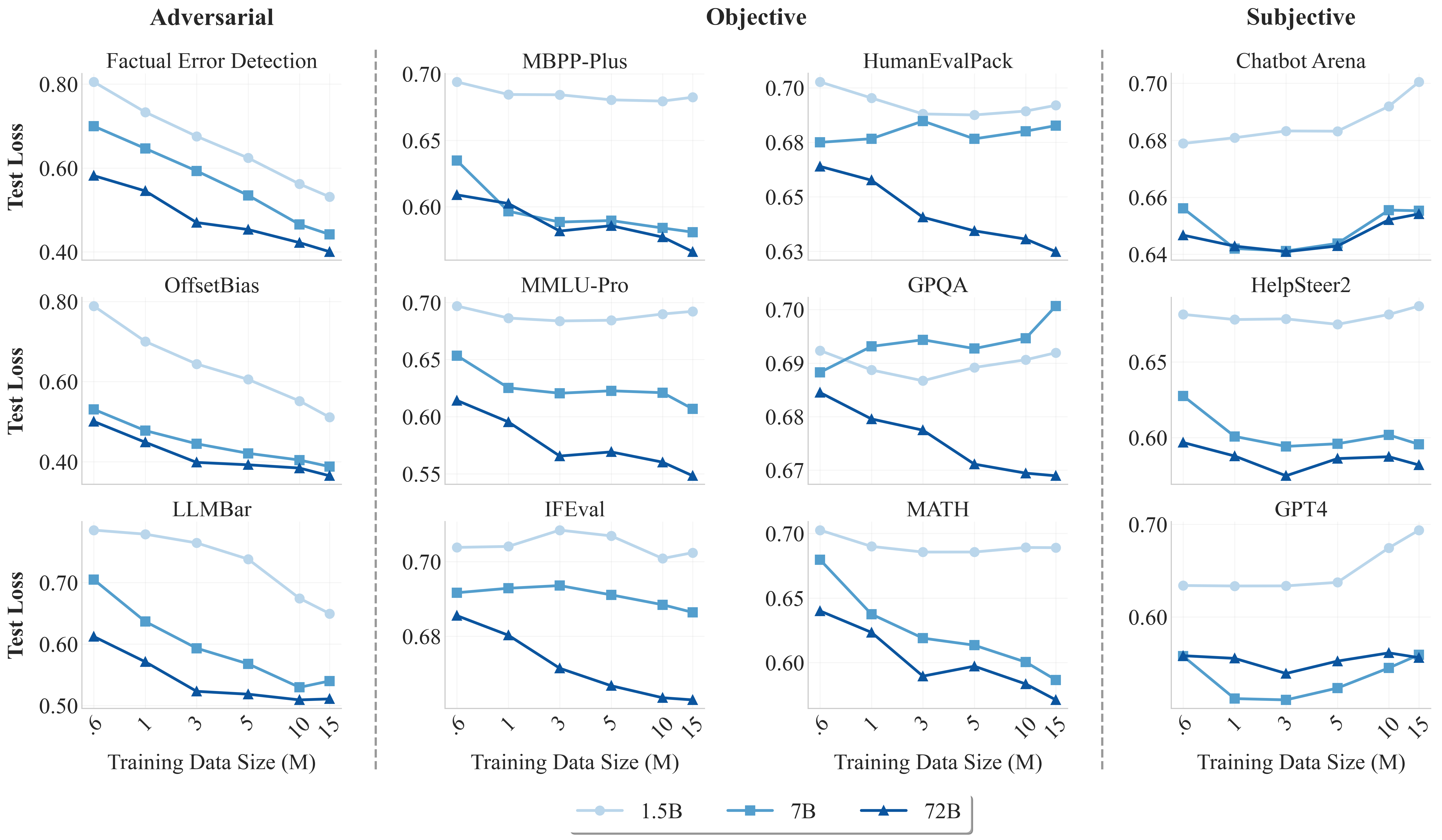}
    \caption{Test loss analysis across domains for 1.5B-72B models with increasing training data. The x-axis is scaled logarithmically (base 2). }
    \label{fig:main-compare-loss}
\end{figure}

\end{abstract}

\section{Introduction}

\citet{Kaplan2020ScalingLF} demonstrates that language modeling performance exhibits scaling laws, where cross-entropy loss scales as a power law with model size, dataset size, and the amount of compute. These laws has achieved remarkable success in next-token prediction tasks, continuously advancing the capabilities of language models~\citep{Bai2023QwenTR, Yang2024Qwen2TR,Yang2024Qwen25TR}. 
We argue that such scaling laws can be extended to preference modeling and thus propose World Preference Modeling (WorldPM), which aims to achieve a unified representation of preferences by leveraging these laws.
A key challenge to this investigation lies in the scalability constraints of manually annotated preference data~\citep{Wang2024HelpSteer2OD,Bai2022ConstitutionalAH,Lee2023RLAIFVR}.

As platforms that facilitate large-scale interaction and opinion sharing, online forums aggregate vast amounts of user preference data through voting mechanisms, substantially advancing research on collective decision making in social sciences and psychology~\citep{easley2010networks, Pal2012ExploringQS, Hu2008CollaborativeFF}. Despite the immense value of these preference signals, they have rarely been effectively leveraged at scale by artificial intelligence. 
In one of the few efforts, \citet{Askell2021AGL} utilized forum data for Preference Model Pre-training (PMP). However, they mainly focus on how PMP improves sample efficiency during the preference fine-tuning process, leaving the scaling properties of preference pre-training itself unexplored.

We first collect preference data from public forums, including StackExchange, Reddit, and Quora. After assessing the quality of different data sources, we find that StackExchange exhibits the best generalization and cross-domain transfer capabilities. Therefore, we select preference data from StackExchange as the representative source for WorldPM. We conduct large-scale preference training with 15 million training samples on language models with parameters ranging from 1.5B to 72B. Subsequently, we evaluate the test loss across various model sizes and training scales using 12 test sets, broadly classified into adversarial, objective, and subjective categories based on the capabilities they assess. The main results in Figure~\ref{fig:main-compare-loss} prove that WorldPM displays distinct scaling patterns across different domains. Specifically: 

(1) \textbf{In adversarial evaluation, test losses demonstrate a power law decrease}, underscoring the model's enhanced ability to detect responses containing intentional errors, as well as those that are well-written but irrelevant or incomplete. This finding suggests that subtle deficiencies commonly plaguing discriminative models~\citep{Park2024OffsetBiasLD,zeng2024llmbar} can be effectively mitigated through large-scale training.

(2) \textbf{The objective evaluation reveal an emergent scaling phenomenon}, where larger models demonstrate a power law reduction in test losses across more benchmarks, whereas smaller models show limited or no such improvements.  These results suggest that WorldPM represents a challenging task that requires larger models to elicit preferences for objective knowledge, pointing to its substantial potential for further advancement.

(3) \textbf{Subjective evaluations show no scaling trends}. 
We analyze potential reasons from the perspective of style preferences (e.g., preference for longer or richer responses). 
While WorldPM becomes more style-neutral as it scales up, some subjective evaluations exhibit style preferences, resulting in lower evaluation performance.
Similar phenomena have been observed in many other studies~\citep{dubois2024length, Park2024OffsetBiasLD, chiang2024chatbot}.

We further explore the potential of WorldPM as a foundation model to facilitate preference fine-tuning. 
Unlike \citet{Askell2021AGL}'s initial work that was limited to small datasets and in-distribution testing, we conduct comprehensive experiments using three human preference datasets (7K, 100K and 800K samples) and evaluate the model on multiple benchmarks.
Results demonstrate that WorldPM significantly improves the generalization capabilities of preference fine-tuning, achieving performance gains exceeding 5\% on many key subtasks. Notably, the improvements are more pronounced when the amount of preference data is limited, with some subtasks showing gains of over 10\%. We also reveal a positive correlation between the scale of WorldPM and the benefits obtained from fine-tuning, indicating scalable advantages as training scale increases.

Finally, we integrate WorldPM into our model training pipeline through preference fine-tuning and Group Relative Policy Optimization (GRPO)~\citep{Shao2024DeepSeekMathPT}. Results show substantial improvements in both in-house and open-sourced alignment evaluations compared to baselines without WorldPM. Notably, our in-house evaluations reveal significant gains ranging from 4\% to 8\%.

\section{Background \& Related Work}

\subsection{Reinforcement Learning from Human Feedback (RLHF)}
\label{sec:background_rlhf}
Reinforcement Learning from Human Feedback (RLHF) has emerged as a critical technique for aligning large language models (LLMs) with human preferences~\citep{christiano2017deep,stiennon2020learning,ouyang2022training, Dubey2024TheL3,Yang2024Qwen2TR, Yang2024Qwen25TR}. The framework typically involves two key stages: \textit{reward modeling} and \textit{reinforcement learning optimization}~\citep{bai2022training,Zheng2023SecretsOR}. As it focuses on human preferences, reward modeling is also known as preference modeling (PM)~\citep{ouyang2022training}. We use both terms interchangeably in this work.

\begin{itemize}
    \item \textbf{Reward Modeling:}  
A reward model (RM) is trained using pairwise comparison data. For each training instance $(x,y_0,y_1)$, we define a preference label $Y \in \{0,1\}$ where $Y=i$ indicates $y_i$ is the \textit{preferred/chosen} response (while the other is \textit{rejected}). The reward model computes scores $r_\theta(x, y_0)$ and $r_\theta(x, y_1)$ for each response, where $r_\theta$ denotes the reward function parameterized by $\theta$. Following the Bradley-Terry (BT) model~\citep{Bradley1952RankAO}, the probability of $y_0$ being preferred over $y_1$ given prompt $x$ is:
\begin{equation}
    P(Y=0|x,y_0,y_1) = \text{sigmoid}(r_\theta(x, y_0) - r_\theta(x, y_1)),
    \label{eq:bt-loss}  
\end{equation}

Typically, the last decoding layer of an LLM is replaced with a linear layer that maps the hidden state of the last token to a scalar value. The training objective minimizes the negative log-likelihood of human preference data, known as the BT loss~\citep{Sun2024RethinkingBM}:
    \begin{equation}
        \mathcal{L}_{BT} = -\mathbb{E}_{(x,y_0,y_1,Y)\sim \mathcal{D}} [\log P(Y|x,y_0,y_1)],
        \label{eq:bt-loss-objective}
    \end{equation}
where $\mathcal{D}$ represents the dataset of preference pairs.

    \item \textbf{Reinforcement Learning:} The trained RM guides policy optimization through RL algorithms like GRPO~\citep{Shao2024DeepSeekMathPT}, which optimizes the policy $\pi_\phi$ by maximizing expected rewards while minimizing KL divergence from a reference policy $\pi_{\text{ref}}$:
        \begin{equation}
            \max_{\phi} \mathbb{E}_{x \sim \mathcal{D}_{\text{prompt}}, \{y_i\}_{i=1}^G \sim \pi_\phi(\cdot|x)} \left[ \frac{1}{G} \sum_{i=1}^G \frac{r_{\theta}(x, y_i) - \mu}{\sigma} - \beta D_{\text{KL}}(\pi_{\phi}(y|x) \| \pi_{\text{ref}}(y|x)) \right],
        \end{equation}
    where $\frac{r_{\theta}(x, y_i) - \mu}{\sigma}$ computes the relative advantage of responses within the group, and $D_{\text{KL}}$ constrains policy optimization from deviating too far from the initial model through KL divergence.

\end{itemize}

\subsection{Best-of-N Sampling as an Alignment Alternative}
Best-of-N (BoN) sampling provides a simple yet effective alternative to RL-based alignment~\citep{xu2024dpo, gao2023scaling}. The method operates as follows:
\begin{enumerate}
    \item \textbf{Responses Sampling:} For a given prompt $x$, generate $N$ candidate responses $\{y_1, ..., y_N\}$ using a policy model (e.g., SFT model).
    \item \textbf{Reward Ranking:} Score all candidates using a reward model $r_\theta(x, y)$.
    \item \textbf{Response Selection:} Output the response with the highest reward score: $y^* = \arg\max_{y_i} r_\theta(x, y_i)$.
\end{enumerate}
BoN decouples alignment from complex RL-based optimization, offering stability and reproducibility by directly leveraging RM rankings. Empirical studies~\citep{gao2023scaling, coste2023reward,ivison2024unpacking,li2023remax} demonstrate that BoN achieves competitive performance with RLHF while avoiding optimization instability.

\subsection{Preference Model Pre-training}
Given the high cost of preference annotation, \citet{Askell2021AGL} propose using large-scale public forum data like StackExchange for Preference Model Pre-training (PMP), making several key findings: (1) PMP improves sample efficiency in small-scale preference fine-tuning; (2) PMP datasets are transferable to different fine-tuning datasets; (3) binary preference modeling outperforms rank modeling.

Building upon these findings, we maintain consistency with their binary preference modeling approach. However, our work differs in several aspects: (1) While they demonstrate PMP's benefits in improving sample efficiency during preference fine-tuning, they do not investigate the inherent scaling properties of large-scale preference training. Our work provides an in-depth analysis of these aspects; (2) We find that PMP datasets not only transfer to different fine-tuning datasets but also generalize directly to various test sets; (3) Due to the scarcity of preference data and evaluation sets at that time, they could only experiment with 5K human preference samples and validate on in-distribution sets. In contrast, we utilize more advanced human preference datasets and conduct comprehensive evaluations across a wide range of benchmarks.

\section{Modeling World Preference}

\label{sec:modeling_world_preference}

\subsection{Experimental Setup}

\subsubsection{Data Collection}
To start modeling world preference, we first collect data from multiple public forums, including StackExchange\footnote{https://stackexchange.com} (a professional Q\&A platform), Reddit\footnote{https://www.reddit.com} (a social news and community discussion platform), and Quora\footnote{https://www.quora.com} (a knowledge sharing and Q\&A community). Each forum contains numerous posts, typically in the form of questions, with responses from different users. Users can upvote or downvote these responses, naturally establishing a relative preference pattern. For each post serving as prompt $x$, we randomly sample two responses with different net votes (upvotes minus downvotes) from its response list to form preference pairs, where we denote the response with higher and lower net votes as $y_w$ and $y_l$ respectively.
Detailed forum data analysis and preference pair sampling strategies (e.g., controlling for vote margin between pairs) are provided in Appendix~\ref{sec:detail_of_settings}. Training examples are presented in Appendix~\ref{sec:case_study}.

\subsubsection{Training Methods}
Our approach to world preference modeling follows the general human preference modeling framework as described in Section~\ref{sec:background_rlhf}. Given a pair of preference samples, we use the preference model to predict their respective rewards and optimize the BT-loss (Eq.~\ref{eq:bt-loss-objective}). For models of different sizes, we maintain consistent hyperparameters with a learning rate of 3e-6 and a batch size of 10K. Ablation studies on learning rate and batch size configurations are provided in Appendix~\ref{sec:args_settings}.

\subsubsection{Evaluation Methods}
To comprehensively evaluate WorldPM, we utilize different domain test sets from multiple RM benchmarks. Given our specific evaluation needs, we do not strictly follow the evaluation protocols provided with these benchmarks. Detailed explanations can be found in Appendix~\ref{sec:evaluation_settings}. This is primarily because: (1) we use BT-Loss (Eq.\ref{eq:bt-loss-objective}) to calculate test performance, which requires original preference pairs (RMB), and (2) we provide detailed style analysis, thus discarding the style control from the RM-Bench benchmark. The benchmarks used in this work are as follows:
\begin{itemize}
\item \textbf{PPE}~\citep{Frick2024HowTE}: This includes evaluations of both subjective and objective parts. The data for the subjective part comes from real user annotations in the Chatbot Arena\footnote{https://lmarena.ai/}. The objective part collects queries from the MMLU-Pro~\citep{Wang2024MMLUProAM}, IFEval~\citep{zhou2023instructionfollowingevaluationlargelanguage}, GPQA~\citep{Rein2023GPQAAG}, MATH~\citep{hendrycksmath2021}, and MBPP-Plus~\citep{Austin2021ProgramSW} datasets, and gathers responses from state-of-the-art models. The correctness of the responses is validated against real answers, thereby forming preference pairs (with \textit{chosen} as correct and \textbf{rejected} as incorrect).
\item \textbf{RMB}~\citep{Zhou2024RMBCB}: This benchmark relies on GPT4 as the primary annotator, with additional human verification processes. It encompasses diverse scenarios and is fundamentally divided into two aspects: helpfulness and harmlessness.
\item \textbf{RM-Bench}~\citep{Liu2024RMBenchBR}: This consists of evaluations in four domains: chat, code, math, and safety. The chat domain evaluation assesses models' ability to identify factual errors by inserting them into responses. Prompts for code and math are sourced from the HumanEvalPack~\citep{muennighoff2023octopack} and MATH~\citep{hendrycksmath2021} datasets, respectively, with validation against real answers. The safety section includes both pseudo-harmful and genuinely harmful questions to evaluate the model's safety assessment capability.
\item \textbf{Reward Bench}~\citep{lambert2024rewardbench}: This includes evaluations in four domains: chat, chat-hard, reasoning, and safety. The chat-hard section, primarily sourced from LLMBar~\citep{zeng2024llmbar}, challenges reward models through the construction of subtly flawed responses designed to mislead evaluation.
\item \textbf{Offset Bias}~\citep{Park2024OffsetBiasLD}: This dataset constructs high-quality but incorrect responses (rejected responses) to challenge reward models, including off-topic responses and responses containing errors.
\item \textbf{HelpSteer2}~\citep{Wang2024HelpSteer2OD}: This dataset, carefully annotated and filtered by trained human annotators, serves dual purposes: while it will be used as a training set in later sections, during this phase it functions as a test set for evaluating WorldPM's subjective performance. 

\end{itemize}

The capabilities tested by the above benchmarks can be broadly classified into three categories: (1) adversarial (identifying flaws in responses, such as constructing irrelevant rejected responses). (2) objective (identifying correct responses for querys with ground-truth answers), and (3) subjective (including human or AI subjective preferences). 

To evaluate WorldPM models' effectiveness in downstream alignment tasks, we employ two benchmarks (Alpaca Eval~\citep{dubois2024length} and Arena Hard~\citep{arenahard2024}) and implement BoN sampling as an alternative to RLHF. These two benchmarks use AI as the subjective evaluator.

\subsubsection{Training Data Source Selection}

\begin{wrapfigure}{R}{0.35\textwidth}
    \vspace{5pt} 
    \includegraphics[width=0.35\textwidth]{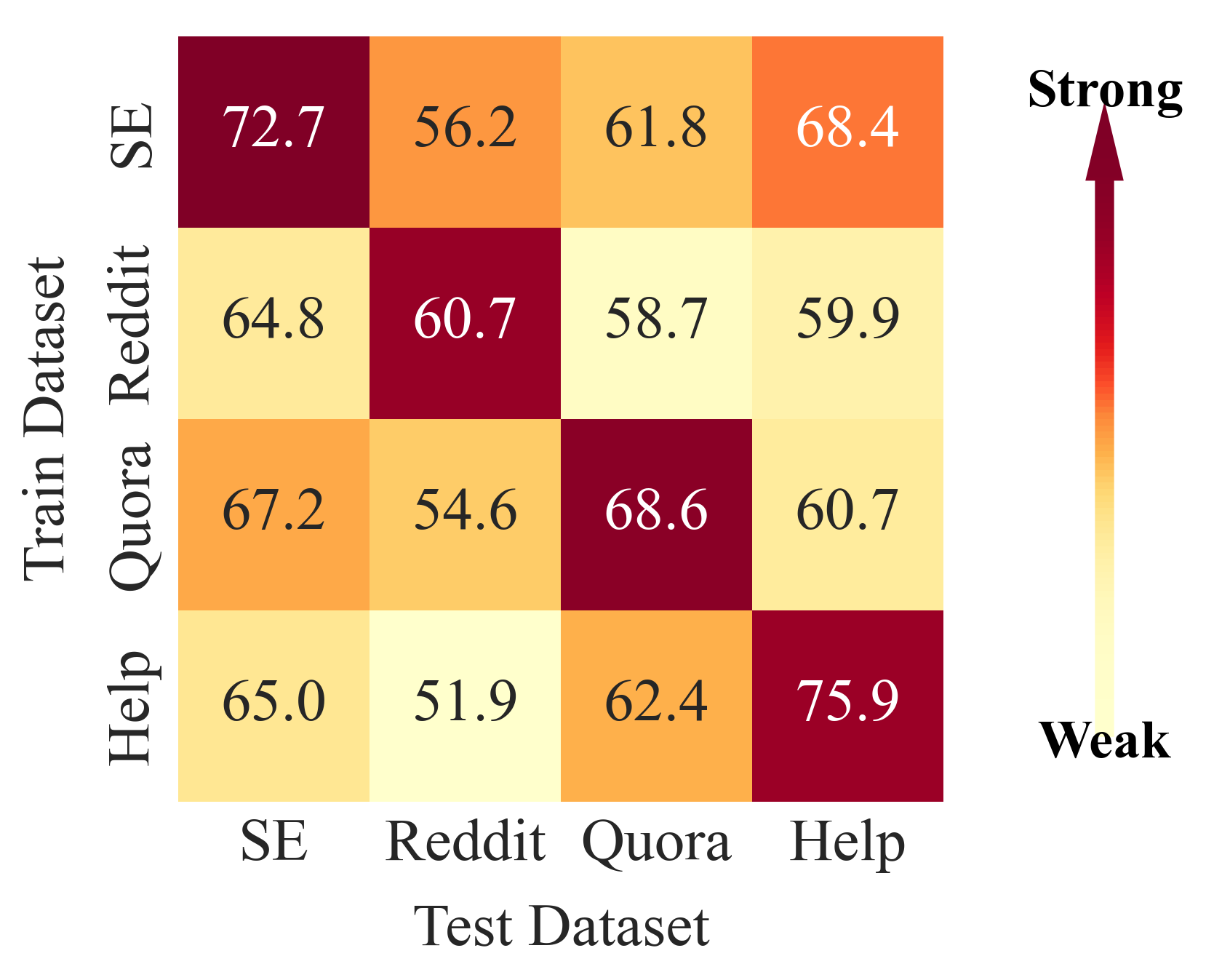}
    \caption{Cross generalization across different data sources including StackExchange, Reddit, Quora, and HelpSteer2, where models trained on one source and predict preferences on the others. The values indicate test accuracy.}
    \vspace{-20pt}  
    \label{fig: cross-generalize}
\end{wrapfigure}

\begin{table}[!htbp]
\centering
\resizebox{0.8\textwidth}{!}{
\begin{tabular}{lccccc}
\toprule
\textbf{Dataset} & \textbf{PPE-Human} & \textbf{PPE-Objective} & \textbf{RMB} & \textbf{RewardBench} & \textbf{RM-Bench} \\
\midrule
Quora & 60.2 & 57.7 & 66.1 & 69.0 & 60.9 \\
Reddit & 55.1 & 59.4 & 59.5 & 77.0 & 60.7 \\
StackExchange & 62.8 & \textbf{62.1} & \textbf{76.7} & \textbf{84.4} & 72.5 \\
\quad  \textit{$\hookrightarrow$ Math SE} & \textbf{62.9} & 62.0 & 75.0 & 83.3 & \textbf{75.0} \\
\midrule
ArmoRM & 60.2 & 64.3 & 72.3 & 89.8 & 75.4 \\
\bottomrule
\end{tabular}}
\caption{Performance comparison of general human preference across different data sources. StackExchange significantly outperforms other sources, approaching or surpassing open-source preference models (ArmoRM-Llama3-8B-v0.1). Math SE, a mathematics-specific board of StackExchange, demonstrates strong general human preference despite its domain-specific nature.}
\label{tab:diff_source_compare}
\end{table}

We collect 800K preference pairs from each forum and train them on Qwen2.5-7B. As shown in Table~\ref{tab:diff_source_compare}, we evaluate them using multiple benchmarks, where PPE-Object is the average of five PPE objective evaluation sets. RMB, RewardBench, and RM-Bench calculate the average of their respective subsets. We also include ArmoRM-Llama3-8B-v0.1\citep{ArmoRM} as a reference. It can be found that \textbf{StackExchange preference data shows the highest quality and demonstrates strong out-of-domain generalization ability, approaching or even surpassing open-source preference models.} Specifically, StackExchange data significantly outperforms Reddit and Quora across all evaluation sets, demonstrating its high-quality. Comparing StackExchange with ArmoRM, we find comparable performance on PPE-Objective and RM-Bench, while even surpassing ArmoRM on PPE-Human and RMB. Given that StackExchange's content is entirely out-of-domain from downstream general preference evaluations, its strong generalization ability is particularly impressive.

To further understand StackExchange's out-of-domain generalization ability, we consider an extreme data source: StackExchange's Math board, which contains only mathematics-related content. Models trained on this board perform similarly to those trained on StackExchange data across various human preference benchmarks, indicating that \textbf{general human preferences can transfer across different domains.}

Besides, we verify StackExchange's generalization ability across different data sources, including HelpSteer2 as a few-human-annotated data source. Models are trained separately on StackExchange, Reddit, Quora, and HelpSteer2, then cross-evaluated on each other's data, as shown in the Figure~\ref{fig: cross-generalize}, colors are normalized per column due to varying dataset difficulties, while values show original accuracies. \textbf{StackExchange shows the best generalization ability among multiple forum data sources and can represent different data sources.}

Given StackExchange's superior quality compared to other sources, we select it as the representative forum data source for modeling world preference.

\subsection{Scaling Trends}
\begin{wrapfigure}{R}{0.4\textwidth}
    \includegraphics[width=1.0\linewidth]{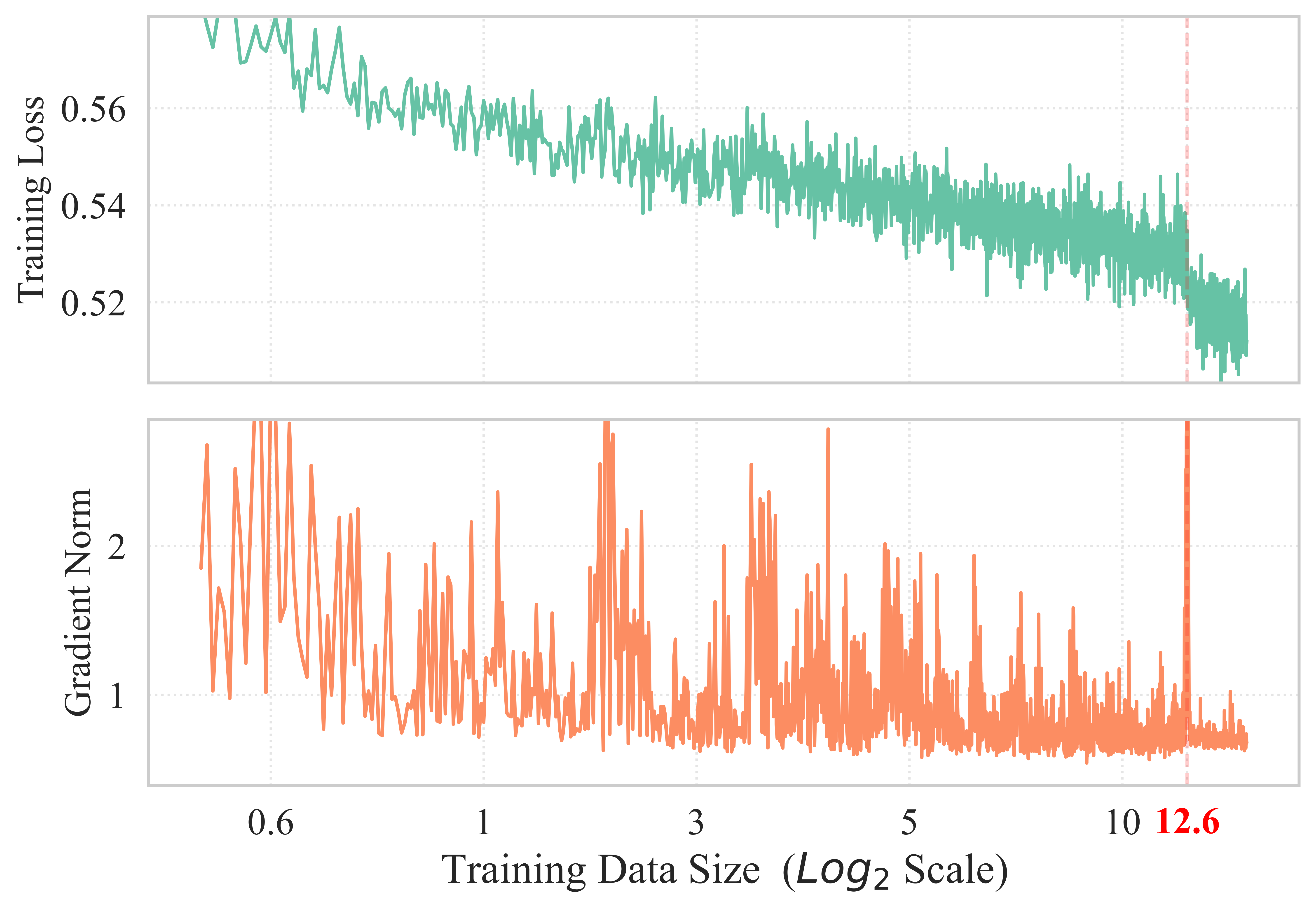}
    \caption{A moment of epiphany occurs during WorldPM training on Qwen 72B, characterized by a sudden drop in loss and a corresponding spike in gradients.}
    \label{fig:emerge_loss}
\end{wrapfigure}

We collect as many posts, responses, and votes as possible from StackExchange to construct preference pairs, yielding approximately 15M pairs. We conduct world preference modeling on Qwen2.5 models~\citep{Yang2024Qwen25TR} ranging from 1.5B to 72B parameters.

During large-scale WorldPM training, we observe a \textbf{moment of epiphany}. As shown in Figure~\ref{fig:emerge_loss}, the training loss exhibits logarithmic decrease with increasing training data volume. All training is completed \textbf{within a single epoch}, without reusing any training samples. At approximately 12.6M training samples, we observe a critical transition: a sudden drop in loss coincides with a sharp spike in gradients that quickly subsides. This synchronized pattern—the abrupt gradient surge indicating the discovery of a steep optimization direction, and the subsequent loss reduction suggesting a successful transition to a better solution space—implies that the model has discovered a more general and effective preference representation, potentially leading to stronger generalization capabilities across different preference scenarios.

Subsequently, we evaluate WorldPM's out-of-domain generalization capabilities across various test sets. We find that WorldPM exhibits different scaling properties across various domains and model sizes, as illustrated in Figure~\ref{fig:main-compare-loss}. The test loss is measured using BT loss (Equation~\ref{eq:bt-loss-objective}). Overall, WorldPM demonstrates scaling trends on test sets with well-defined answers (Objective and Adversarial), where more challenging test sets require larger models to achieve scaling benefits. However, no clear scaling trends are observed on test sets with open-ended answers (Subjective).

\textbf{We observe a power law decrease in test losses in adversarial evaluation.} 
To construct adversarial evaluations, we utilized three test sets: factual error detection (from RM-Bench chat), Offsetbias, and LLMBar (from RewardBench chat-hard). These test sets challenge reward models' robustness by either inserting factual errors into responses or constructing off-topic/incomplete responses. 
Notably, models of varying sizes exhibited a power-law decrease in test losses across all three test sets. Multiple studies~\citep{Park2024OffsetBiasLD, zeng2024llmbar} have identified inherent biases and vulnerabilities in reward models that can be exploited by RLHF for reward hacking (e.g., assigning high scores to superficially good but unfaithful responses). Our findings suggest that such vulnerabilities may stem from insufficient training data. As training data expands, reward models demonstrate increasingly stronger capability to identify flawed responses. 

\textbf{We observe an emergent phenomenon in objective metrics, where larger models demonstrate power law decrease in test losses across more benchmarks.} We evaluate multiple aspects using six objective test sets: coding (MBPP-Plus, HumanEvalPack), mathematics (MATH), knowledge-based QA (MMLU-Pro, GPQA), and instruction following (IFEval). MATH and HumanEvalPack are sourced from RM-Bench's code and math domains, while others are from PPE. We find that the 1.5B model fails to generalize on any objective test set. The 7B model shows weak power law decrease on MBPP-Plus and IFEval, but exhibits increasing losses on more challenging benchmarks like HumanEvalPack and GPQA. In contrast, the 72B model exhibits consistent scaling trends across all aspects. These results indicate that preference modeling is an inherently challenging task, where \textbf{certain capabilities emerge only at larger model scales} - exemplifying the emergence phenomenon in large language models \citep{Wei2022EmergentAO}.

\textbf{Test losses in subjective evaluations quickly converge or even show an increasing trend.} We collected annotations from three distinct sources:
\begin{enumerate}
    \item \textbf{Crowdsourced annotations} (ChatBot Arena) from PPE's Human Preference test set, which contains authentic human preference selections from ChatBot Arena without strict control over the annotation process;
    \item \textbf{Expert annotations} from HelpSteer2, following specific guidelines and employing multiple annotators with consensus-based quality control;
    \item \textbf{GPT4 annotations} from RMB's Helpfulness evaluation, which implements a set of human preference guidelines for GPT4-based annotation - a strategy widely adopted in alignment evaluations, including Alpaca Eval and Arena Hard.
\end{enumerate}

We find that models from 1.5B to 72B do not show decreasing test losses with increased training data. From the model size perspective, the 1.5B model shows significant performance differences from the other two, while there is no notable distinction between 7B and 72B models. These phenomena stand in stark contrast to results from other evaluations. 

\textbf{We hypothesize that this might be due to conflicts between WorldPM and subjective evaluation in certain dimensions.} Subjective evaluation encompasses a rich variety of dimensions. When determining which response is better, humans or AI may consider multiple aspects such as usefulness, relevance, and conciseness~\citep{arenahard2024}. Different individuals may have varying preferences - some may favor concise responses while others prefer verbose ones.
Furthermore, within the same dimension, different people may have different interpretations; for example, given the same pair of responses, some people might find response A more useful while others might consider response B more helpful. When these noisy preference annotations are used as test sets for evaluation, they can introduce uncertainty into the assessment results~\citep{dubois2024length, Park2024OffsetBiasLD}. For instance, when a preference model rewards brevity while a subjective evaluation favors verbosity, the final evaluation results may appear poor, even though the preference model performs well in other dimensions such as usefulness. Appendix~\ref{sec:case_study} presents several cases of questionable human annotations from ChatBot Arena.

\subsection{Style Impact Analysis}

\subsubsection{Style-Content Separation in Evaluation}

\begin{figure}[b]
    \centering
    \includegraphics[width=\linewidth]{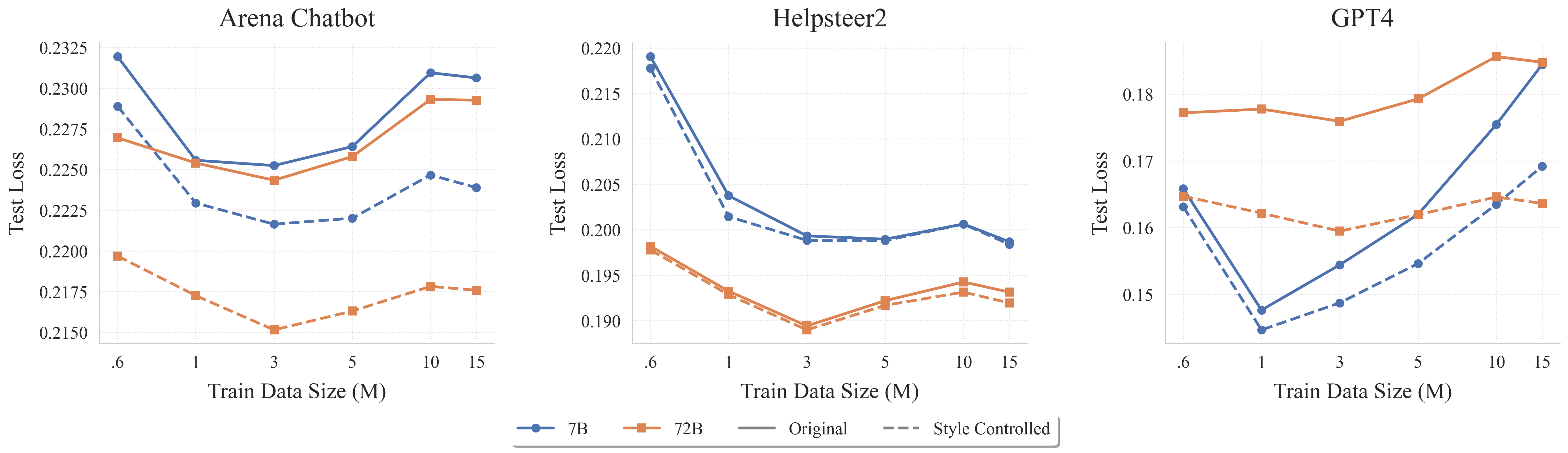}
    \caption{Comparison of test loss on subjective evaluation sets across annotation sources, with and without style control. HelpSteer2's expert annotations show minimal impact from style control, while crowdsourced annotations (ChatBot Arena) and AI annotations (GPT4) show substantial variations. The gap between controlled and uncontrolled conditions grows with training data and model size, reflecting WorldPM's reduced style preference.}
    \label{fig:Compare-Style-Control}
\end{figure}

To investigate challenges in subjective evaluation, we focus on style preference as a reliably quantifiable aspect, which has been shown to influence LLM evaluation in multiple studies~\citep{Zheng2023JudgingLW, dubois2024length,chiang2024chatbot,Feuer2024StyleOS}.

Inspired by \cite{chiang2024chatbot}, we propose to separate style evaluation and content evaluation during the  preference model assessment. For a given preference pair \( (x, y_0, y_1) \), we define the reward model scores as \( r_\theta(x, y_0) \) and \( r_\theta(x, y_1) \). Consequently, the difference in model scores is given by \( D(x, y_0, y_1) = r_\theta(x, y_0) - r_\theta(x, y_1) \). Let \( Z \in \mathbb{R}^S \) denote the style features, where \( S \) is the number of style features. Consistent with \cite{chiang2024chatbot}, we quantify response's style using four features: token length, number of markdown lists, headers, and bold elements.

The style difference for the preference pair can be defined as:
\[
Z(x, y_0, y_1) = \text{normalize}\left(\frac{Z(y_0) - Z(y_1)}{Z(y_0) + Z(y_1)}\right).
\]

We combine the score difference \( D \) and the style difference \( Z \) linearly to form the final evaluation metric:
\begin{equation}
\label{eq:final_score}
R = D^T \alpha + Z^T \beta,
\end{equation}

where \( \alpha \) and \( \beta \) are weight coefficients that adjust the influence of score differences and style differences in the final evaluation.
The optimal values of \( \alpha \) and \( \beta \) are obtained by minimizing Equation~\ref{eq:style_control_loss} through the linear regression algorithm. This formula maintains consistency with the training objective in Equation~\ref{eq:bt-loss-objective}.

\begin{equation}
\label{eq:style_control_loss}
\hat{\alpha}, \hat{\beta} = \arg \min_{\beta \in \mathbb{R}, \gamma \in \mathbb{R}^S} \frac{1}{n} \sum_{i=1}^{n} -(Y_i \log(\text{sigmoid}(R_i)) + (1-Y_i) \log(1-\text{sigmoid}(R_i))),
\end{equation}

where \( R_i \) is the final reward difference for each preference pair. 
From an intuitive perspective, explicitly modeling stylistic factors during evaluation maximizes the stylistic gain in evaluation outcomes across different models, thus effectively eliminating stylistic influences when conducting comparisons across various models.

We compare test losses with and without style control across subjective test sets (style control uses \( R_i \) as the final score, while no control sets $\beta=0$ in Equation~\ref{eq:final_score}\footnote{This differs from original test loss calculation as $\alpha$ is still optimized for controlled comparison, yielding lower values}), as shown in Figure~\ref{fig:Compare-Style-Control}. Expert annotations (HelpSteer2) results remain stable - potentially due to strict quality control in its construction. However, test sets from crowdsourced annotations (ChatBot Arena) and AI annotations (GPT4) show significant variations. This suggests that \textbf{without careful annotation, subjective evaluations are highly sensitive to style factors, while well-controlled test sets remain stable.}

Examining the gap between controlled and uncontrolled accuracies (shown as the space between dashed and solid lines), we observe that this gap widens with increased training and model scale, consistent with WorldPM's gradual reduction in style preference, as discussed in the following sections. After style control, the 72B model outperforms the 7B model (though still slightly underperforming on GPT4 evaluations, possibly due to additional uncontrolled preferences). This validates the benefits of model scaling in subjective tasks, although the lack of improvement with training scale remains an open question - we suspect that world preferences may still conflict with certain preference in subjective evaluations.

Further analysis of stylistic factors in evaluation is presented in the Appendix~\ref{sec:detail_of_style_control}, encompassing impact across different domains and feature andablation studies. Although stylistic influence varies across different domain test sets, it does not alter the primary trends observed.

\subsubsection{Style Effects on Training Dynamics}
\begin{figure}[htbp]
    \centering
    \begin{subfigure}[t]{0.48\textwidth}
        \includegraphics[width=\linewidth]{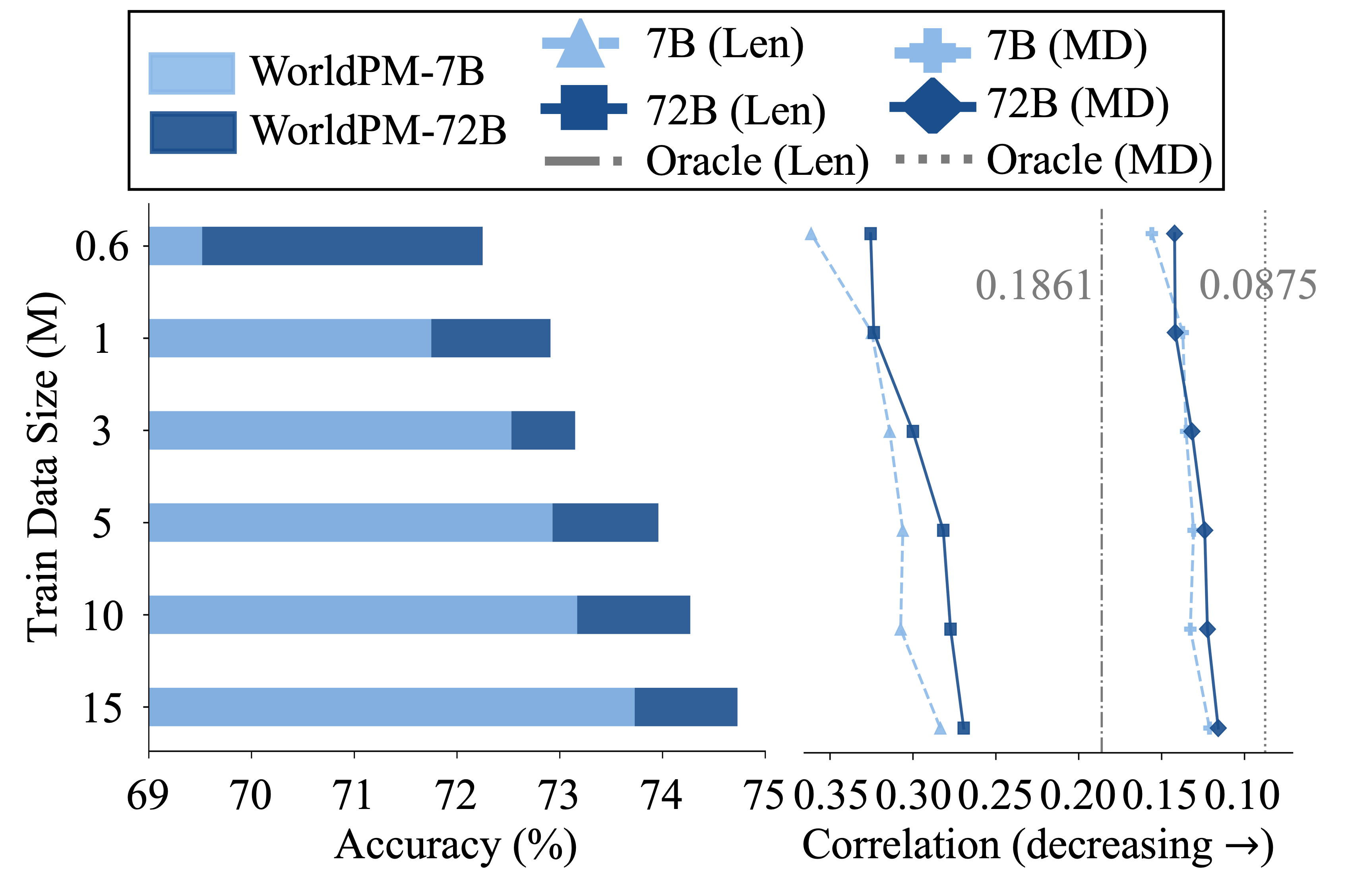}
        \caption{The left panel shows continuous performance improvements, while the right panel demonstrates decreasing correlation between model predictions and stylistic features as model size and training scale increase. However, this correlation remains higher than that (Oracle) between original labels and style, suggesting over-reliance on stylistic features.}
        \label{fig:style-corr}
    \end{subfigure}
    \hfill 
    \begin{subfigure}[t]{0.45\textwidth}
        \includegraphics[width=0.85\linewidth]{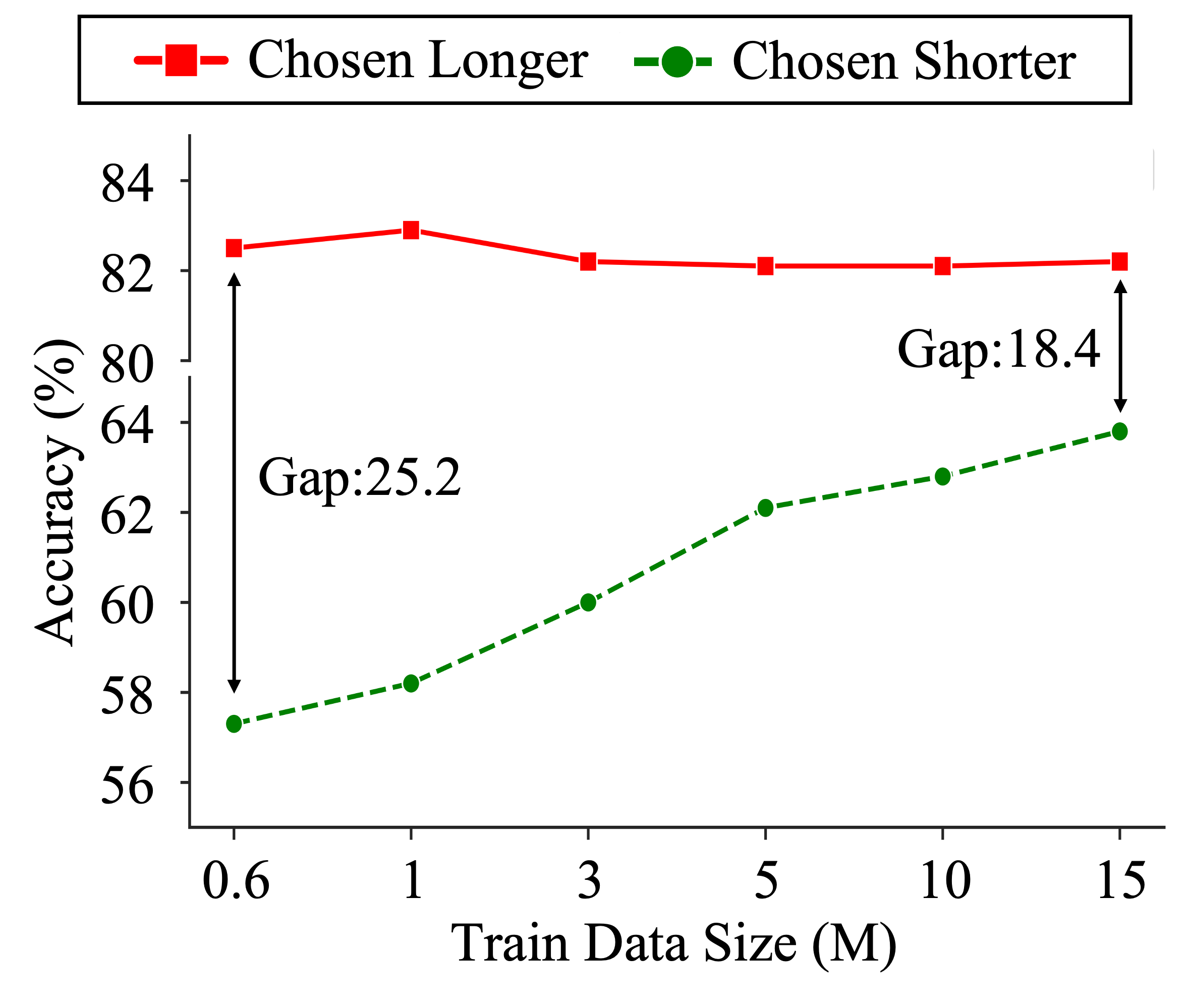}
        \caption{An asymmetric learning dynamic in style preference mitigation. The model rapidly fits majority-style (Chosen Longer) data while gradually incorporating minority-style instances (Chosen Shorter), maintaining performance on the former (Chosen Longer) throughout the training process.}
        \label{fig:two-part}
    \end{subfigure}
    \label{fig:combined}
\end{figure}

Our previous analysis revealed style preference in subjective evaluation, showing increasing distortion in evaluation results as training scale and model size expand. To investigate the underlying causes of this amplified distortion, we analyze the influence of stylistic factors during WorldPM's training process.

\textbf{We find that models initially over-rely on stylistic features; however, this stylistic bias is reduced as training data and model size increase.} 
To measure the extent to which WorldPM predictions are influenced by stylistic features, given a preference pair \((x, y_0, y_1)\), we define \(S(x, y_0, y_1)\) to indicate which response has better stylistic features. \(P(x, y_0, y_1)\) represents the model's prediction of which response is preferable, while \(G(x, y_0, y_1)\) is the human preference label. We use the Phi coefficient ($\phi$) to determine their correlation, which characterizes the correlation between two binary variables~\citep{cramer1946mathematical}. Specifically, $\phi(P,S)$ measures how strongly model predictions are influenced by stylistic features, while $\phi(G,S)$ reflects the relationship between human preferences and style. $\phi$ is calculated as:
\begin{equation}
\phi(i,j) = \frac{n_{11}n_{00} - n_{10}n_{01}}{\sqrt{(n_{1\cdot})(n_{0\cdot})(n_{\cdot1})(n_{\cdot0})}}
\label{eq:phi}
\end{equation}
where \(n_{ij}\) denotes the count of preference pairs where the first variable is $i$ and the second is $j$, with $i,j \in {0,1}$. The dot notation \(n_{i\cdot}\) or \(n_{\cdot j}\) represents the marginal sum over the corresponding index.

As shown in Figure~\ref{fig:style-corr}, we conduct style preference analysis on the in-distribution validation set. The left panel demonstrates improving valid accuracy with increased training and model scale, indicating enhanced model performance. For analyzing stylistic influence, we consider two factors: \(S_{length}\) (indicating which response is longer) and \(S_{markdown}\) (indicating which response contains more markdown markers). The right panel shows that with larger training data and model sizes, the correlation between model predictions \(P\) and these stylistic features \(S\) tended to decline, suggesting the model's predictions became gradually less dependent on stylistic features.

The vertical lines represent the correlation between human preference labels \(G\) and style \(S\). While human preferences show positive correlation with stylistic features, this correlation is consistently lower than that between model predictions and style, which suggests that the model has over-relied on stylistic features for predictions. This observation aligns with \citep{Geirhos2020ShortcutLI,Holtzman2021SurfaceFC,Zhang2016UnderstandingDL}: when training is insufficient or the model's capabilities are limited, it tends to prioritize surface-level, easily discernible features for making predictions.

Furthermore, we observe that both human labels and model predictions show stronger correlations with text length compared to markdown marker frequency, confirming previous findings\citep{chiang2024chatbot} that markdown usage serves as a secondary stylistic feature.

\textbf{The model quickly fits majority-style data while maintaining long-term memory, spending most training time gradually learning from minority-style instances.} We categorize the validation set into two groups based on length, as it represents the primary stylistic factor: one group where the chosen responses are longer than the rejected ones (Chosen Longer), and the other where the opposite is true (Chosen Shorter). As shown in Figure~\ref{fig:two-part}, we find that the model could classify Chosen Longer data with an 82\% accuracy rate at a very early stage, indicating that it easily learns to distinguish this portion of the data. In contrast, the accuracy for Chosen Shorter data started at only 57\%. As training progressed, the accuracy for Chosen Longer data remained stable overall, whereas the accuracy for Chosen Shorter data exhibited a trend of logarithmic growth with training volume. This reveals the general learning process of WorldPM: the model first captures preference patterns exhibited by the majority of the dataset. However, the remaining data demonstrates contrary characteristics, forcing the model to discover underlying representations that can simultaneously model both majority and minority cases to further reduce loss, thereby overcoming initial biases.

\subsection{Alignment Performance}

\begin{figure}[htbp]
    \centering
    \includegraphics[width=\linewidth]{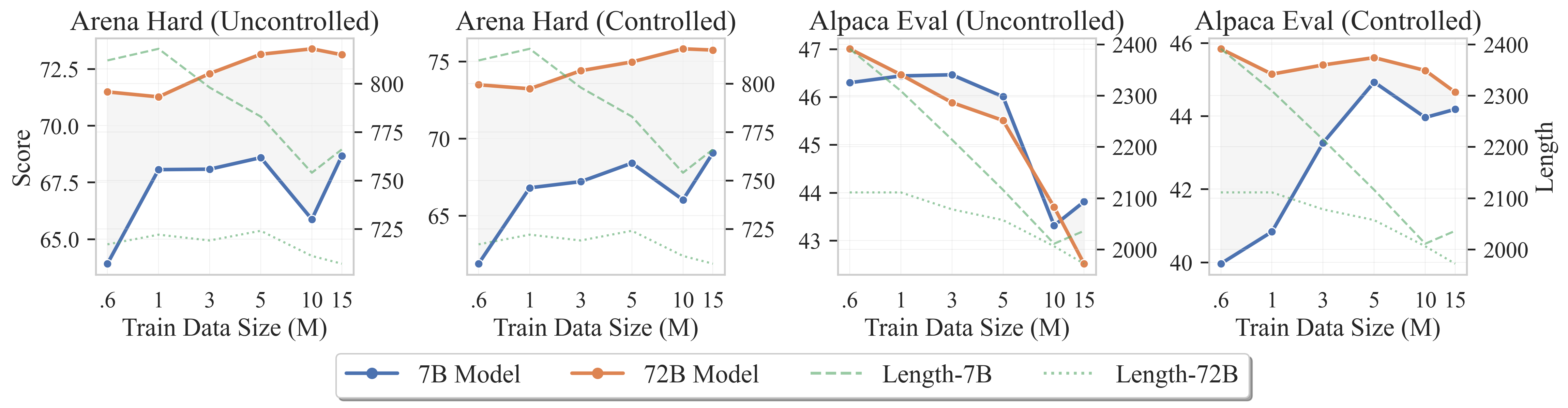}
    \caption{Analysis of alignment performance across WorldPM training scales. Both Alpaca Eval and Arena Hard implement distinct style control mechanisms to mitigate style preference in AI-based subjective evaluation. The figure demonstrates performance under both controlled and uncontrolled conditions. Arena Hard exhibits stable performance trends across control conditions, with 72B consistently superior to 7B and optimal performance achieved at larger training scales. However, Alpaca Eval shows substantial sensitivity to style control, with evaluation scores highly correlated with response length when style control is absent.}
    \label{fig:WorldPM-align}
\end{figure}

To further understand WorldPM's performance in subjective aspects, we evaluate its preference modeling capability indirectly by using PM for BoN sampling to align language models, assessing their performance on Alpaca Eval and Arena Hard. Specifically, for queries from both benchmarks, we generate 256 samples from Qwen2.5-7B-Instruct and let different checkpoints of 7B and 72B WorldPM models select the best response for evaluation. The results are shown in Figure~\ref{fig:WorldPM-align}, presenting both the original scores and scores after applying their respective style control strategies.

We observe that the average length of responses selected by WorldPM gradually decreases across both benchmarks, with 72B generating shorter responses than 7B, consistent with our previous observations. Arena Hard shows stable trends before and after style control, with 72B significantly outperforming 7B, further validating that WorldPM's subjective performance improves with model size. For Alpaca Eval, without style control, performance strongly correlates with length, making it difficult to distinguish between 72B and 7B performance. This again echoes our previous observations - only with style control do we observe clear performance differences between them. Further examining their optimal training scales, we find that the 72B model achieves optimal performance on Alpaca Eval at 0.6M; direct RM evaluation on three test sets shows performance saturation around 3M; in contrast, Arena Hard performance continues to improve until 10M.

Comparing the prompts of these two evaluators (Figure~\ref{fig:evaluator_prompt_arena_hard} and Figure~\ref{fig:evaluator_prompt_alpaca_eval}), we find that Arena Hard explicitly requires answers to be helpful, relevant, and concise (with the conciseness requirement aligning with WorldPM). In contrast, Alpaca Eval only asks models to select the best output from a human perspective and lacks Arena Hard's Chain-of-Thought (COT) process. Consequently, without length control, Alpaca Eval results are strongly correlated with response length. This explains our observation that as training scale increases, performance on Arena Hard continues to improve over a longer period, while performance on Alpaca Eval reaches optimal levels early on.

This comparison highlights the challenges in subjective evaluation. Subjective evaluation encompasses multiple assessment dimensions, such as usefulness, relevance, and conciseness. \textbf{As training scale increases, WorldPM may improve in aspects like conciseness, but some subjective evaluations (like Alpaca Eval) might prefer complex and detailed responses, which conflicts with the goal of conciseness}. We can further speculate that there might be more dimensions where WorldPM's scaling direction diverges from evaluation criteria, leading to continuous increases in losses across these evaluation dimensions, resulting in what appears to be overfitting in subjective evaluations.

\section{Preference Fine-Tuning based on WorldPM }

The previous experiments comprehensively verified the trends observed in various out-of-domain evaluations as the scale and size of training data expanded. Although the subjective evaluations have stagnated, continuous improvements in other aspects suggest that the model has learned more universal representations from WorldPM. Can these representations be utilized during the preference fine-tuning phase? To figure out that, we further validate during fine-tuning based on WorldPM.

\subsection{Experimental Setup}

To evaluate WorldPM as a foundation for preference fine-tuning, we conduct comprehensive experiments using several open-source human preference datasets. These include HelpSteer2~\citep{Wang2024HelpSteer2OD}, UltraFeedback~\citep{cui2023ultrafeedback}, and pair\_data\_v2 800K wsafety\footnote{\href{https://huggingface.co/datasets/RLHFlow/pair_data_v2_80K_wsafety\%7D}{https://huggingface.co/datasets/RLHFlow/pair\_data\_v2\_80K\_wsafety}} from RLHFlow (hereafter referred to as RLHFlow). HelpSteer2 comprises approximately 7K preference comparisons across five dimensions. Since we consider only a single reward, we selected helpfulness scores as the preference labels. UltraFeedback contains four responses per question; we extracted two groups of responses per question (ensuring no overlap between responses), yielding approximately 100K preference pairs. RLHFlow encompasses approximately 800K data points. These three datasets, representing different data scales, enable us to understand the effectiveness boundary of WorldPM. Given HelpSteer2's smaller size, we set its batch size to 128, while maintaining 512 for the other two datasets. All datasets underwent training for a minimum of two epochs, with the final model selected based on the minimum loss achieved on the same distribution validation set.

We comprehensively validate the effectiveness of WorldPM in preference model fine-tuning through two approaches. First, we evaluate various RM benchmark metrics. Second, considering that the primary objective of the preference model is to align language model outputs with human preferences, we employ best-of-N sampling for alignment evaluation. The alignment performance is assessed on two benchmarks - Arena Hard and Alpaca Eval - following our previously described evaluation protocol. Specifically, we generate 64 samples from Qwen2.5-7B-Instruct and Qwen2.5-72B-Instruct for the 7B and 72B models.

\definecolor{up4}{HTML}{87CEEB}     
\definecolor{down4}{HTML}{FFB6C1}

\begin{table}[!htbp]
\centering
\resizebox{0.92\textwidth}{!}{
\begin{tabular}{lcccccc}
\toprule
\multirow{2}{*}{\centering Metrics} & \multicolumn{2}{c}{Helpsteer2} & \multicolumn{2}{c}{UltraFeedback} & \multicolumn{2}{c}{RLHFlow} \\
\cmidrule(lr){2-3} \cmidrule(lr){4-5} \cmidrule(lr){6-7}
 & w/o WorldPM & w/ WorldPM & w/o WorldPM & w/ WorldPM & w/o WorldPM & w/ WorldPM \\
\midrule
\multicolumn{7}{c}{\textbf{Subjective Evaluation}} \\
\midrule
PPE-Human & 63.32 & 64.93 \textcolor{up4}{($\uparrow$2.54\%)} & 63.22 & 63.86 \textcolor{up4}{($\uparrow$1.01\%)} & 66.18 & 66.78 \textcolor{up4}{($\uparrow$0.91\%)} \\
RMB-Helpful & 74.33 & 77.02 \textcolor{up4}{($\uparrow$3.61\%)} & 78.37 & 78.82 \textcolor{up4}{($\uparrow$0.58\%)} & 78.66 & 79.27 \textcolor{up4}{($\uparrow$0.78\%)} \\
RB-Chat & 97.21 & 97.07 \textcolor{down4}{($\downarrow$0.14\%)} & 97.49 & 97.21 \textcolor{down4}{($\downarrow$0.29\%)} & 98.04 & 97.77 \textcolor{down4}{($\downarrow$0.28\%)} \\
\midrule
\multicolumn{7}{c}{\textbf{Objective Evaluation}} \\
\midrule
PPE-MBPP & 72.24 & 73.55 \textcolor{up4}{($\uparrow$1.82\%)} & 69.77 & 73.55 \textcolor{up4}{($\uparrow$5.42\%)} & 75.06 & 76.83 \textcolor{up4}{($\uparrow$2.37\%)} \\
PPE-GPQA & 56.27 & 56.70 \textcolor{up4}{($\uparrow$0.76\%)} & 56.19 & 55.43 \textcolor{down4}{($\downarrow$1.36\%)} & 57.34 & 57.60 \textcolor{up4}{($\uparrow$0.44\%)} \\
PPE-IFEval & 58.81 & 57.34 \textcolor{down4}{($\downarrow$2.49\%)} & 57.42 & 57.03 \textcolor{down4}{($\downarrow$0.68\%)} & 62.09 & 62.38 \textcolor{up4}{($\uparrow$0.47\%)} \\
PPE-MATH & 71.04 & 71.64 \textcolor{up4}{($\uparrow$0.85\%)} & 71.76 & 74.36 \textcolor{up4}{($\uparrow$3.62\%)} & 78.98 & 80.16 \textcolor{up4}{($\uparrow$1.48\%)} \\
PPE-MMLU & 63.55 & 66.52 \textcolor{up4}{($\uparrow$4.67\%)} & 65.64 & 66.66 \textcolor{up4}{($\uparrow$1.55\%)} & 67.97 & 70.27 \textcolor{up4}{($\uparrow$3.39\%)} \\
RM-B-Code & 57.68 & 60.96 \textcolor{up4}{($\uparrow$5.70\%)} & 59.21 & 62.94 \textcolor{up4}{($\uparrow$6.30\%)} & 65.35 & 67.76 \textcolor{up4}{($\uparrow$3.69\%)} \\
RM-B-Math & 70.70 & 75.99 \textcolor{up4}{($\uparrow$7.49\%)} & 72.12 & 75.71 \textcolor{up4}{($\uparrow$4.98\%)} & 76.56 & 80.43 \textcolor{up4}{($\uparrow$5.06\%)} \\
RB-Reason & 88.02 & 96.44 \textcolor{up4}{($\uparrow$9.57\%)} & 92.35 & 95.39 \textcolor{up4}{($\uparrow$3.29\%)} & 95.91 & 96.82 \textcolor{up4}{($\uparrow$0.95\%)} \\
\midrule
\multicolumn{7}{c}{\textbf{Adversarial Evaluation}} \\
\midrule
RB-Chat-Hard & 81.25 & 83.99 \textcolor{up4}{($\uparrow$3.37\%)} & 77.52 & 79.82 \textcolor{up4}{($\uparrow$2.97\%)} & 81.47 & 82.57 \textcolor{up4}{($\uparrow$1.35\%)} \\
RM-B-Chat & 78.29 & 82.17 \textcolor{up4}{($\uparrow$4.95\%)} & 80.23 & 81.01 \textcolor{up4}{($\uparrow$0.97\%)} & 82.17 & 82.56 \textcolor{up4}{($\uparrow$0.47\%)} \\
OffsetBias & 78.87 & 86.60 \textcolor{up4}{($\uparrow$9.80\%)} & 81.68 & 82.94 \textcolor{up4}{($\uparrow$1.55\%)} & 83.43 & 85.69 \textcolor{up4}{($\uparrow$2.71\%)} \\
\midrule
\multicolumn{7}{c}{\textbf{Safety Evaluation}} \\
\midrule
RMB-Harmless & 72.73 & 73.99 \textcolor{up4}{($\uparrow$1.72\%)} & 69.74 & 70.92 \textcolor{up4}{($\uparrow$1.69\%)} & 62.86 & 63.79 \textcolor{up4}{($\uparrow$1.48\%)} \\
RB-Safety & 90.68 & 94.53 \textcolor{up4}{($\uparrow$4.25\%)} & 91.76 & 89.46 \textcolor{down4}{($\downarrow$2.50\%)} & 90.88 & 92.03 \textcolor{up4}{($\uparrow$1.26\%)} \\
RM-B-Safety-R & 90.45 & 93.95 \textcolor{up4}{($\uparrow$3.87\%)} & 96.50 & 96.82 \textcolor{up4}{($\uparrow$0.33\%)} & 92.99 & 96.18 \textcolor{up4}{($\uparrow$3.42\%)} \\
RM-B-Safety-F & 99.65 & 99.30 \textcolor{down4}{($\downarrow$0.35\%)} & 98.94 & 98.94 (0.00\%) & 98.59 & 99.12 \textcolor{up4}{($\uparrow$0.54\%)} \\
\midrule
\multicolumn{7}{c}{\textbf{Alignment Evaluation}} \\
\midrule
Arena Hard & 65.88 & 70.27 \textcolor{up4}{($\uparrow$6.66\%)} & 67.93 & 72.12 \textcolor{up4}{($\uparrow$6.17\%)} & 72.41 & 74.62 \textcolor{up4}{($\uparrow$3.05\%)} \\
Alpaca Eval & 43.73 & 46.29 \textcolor{up4}{($\uparrow$5.84\%)} & 46.16 & 46.35 \textcolor{up4}{($\uparrow$0.40\%)} & 53.24 & 50.97 \textcolor{down4}{($\downarrow$4.27\%)} \\
Arena Hard-SC & 65.07 & 70.22 \textcolor{up4}{($\uparrow$7.91\%)} & 66.19 & 70.36 \textcolor{up4}{($\uparrow$6.30\%)} & 72.77 & 74.29 \textcolor{up4}{($\uparrow$2.09\%)} \\
Alpaca Eval-LC & 39.98 & 42.14 \textcolor{up4}{($\uparrow$5.41\%)} & 44.46 & 46.13 \textcolor{up4}{($\uparrow$3.76\%)} & 46.24 & 46.32 \textcolor{up4}{($\uparrow$0.17\%)} \\
\bottomrule
\end{tabular}}
\caption{Effect of WorldPM initialization on 7B model's PM fine-tuning performance across different evaluation categories. Abbreviations: RB - RewardBench, RM-B - RM-Bench, Safety-R/F - Safety-Response/Refuse.}
\label{tab:fintune-7b}
\end{table}

\begin{table}[!htbp]
\centering
\small
\resizebox{0.92\textwidth}{!}{
\begin{tabular}{lcccccc}
\toprule
\multirow{2}{*}{\centering Metrics} & \multicolumn{2}{c}{Helpsteer2} & \multicolumn{2}{c}{UltraFeedback} & \multicolumn{2}{c}{RLHFlow} \\
\cmidrule(lr){2-3} \cmidrule(lr){4-5} \cmidrule(lr){6-7}
 & w/o WorldPM & w/ WorldPM & w/o WorldPM & w/ WorldPM & w/o WorldPM & w/ WorldPM \\
\midrule
\multicolumn{7}{c}{\textbf{Subjective Evaluation}} \\
\midrule
PPE-Human & 64.76 & 66.68 \textcolor{up4}{($\uparrow$2.97\%)} & 65.12 & 66.94 \textcolor{up4}{($\uparrow$2.80\%)} & 68.14 & 68.93 \textcolor{up4}{($\uparrow$1.16\%)} \\
RMB-Helpful & 77.76 & 78.53 \textcolor{up4}{($\uparrow$1.00\%)} & 79.80 & 81.73 \textcolor{up4}{($\uparrow$2.41\%)} & 79.26 & 80.89 \textcolor{up4}{($\uparrow$2.05\%)} \\
RB-Chat & 98.04 & 98.88 \textcolor{up4}{($\uparrow$0.85\%)} & 97.63 & 96.37 \textcolor{down4}{($\downarrow$1.29\%)} & 98.46 & 97.91 \textcolor{down4}{($\downarrow$0.57\%)} \\
\midrule
\multicolumn{7}{c}{\textbf{Objective Evaluation}} \\
\midrule
PPE-MBPP & 70.50 & 78.76 \textcolor{up4}{($\uparrow$11.72\%)} & 75.02 & 79.46 \textcolor{up4}{($\uparrow$5.92\%)} & 82.28 & 82.39 \textcolor{up4}{($\uparrow$0.14\%)} \\
PPE-GPQA & 57.66 & 59.63 \textcolor{up4}{($\uparrow$3.42\%)} & 59.22 & 61.19 \textcolor{up4}{($\uparrow$3.33\%)} & 62.52 & 63.75 \textcolor{up4}{($\uparrow$1.97\%)} \\
PPE-IFEval & 59.24 & 63.63 \textcolor{up4}{($\uparrow$7.42\%)} & 60.57 & 64.34 \textcolor{up4}{($\uparrow$6.22\%)} & 64.26 & 65.39 \textcolor{up4}{($\uparrow$1.76\%)} \\
PPE-MATH & 69.94 & 77.17 \textcolor{up4}{($\uparrow$10.33\%)} & 72.77 & 78.16 \textcolor{up4}{($\uparrow$7.41\%)} & 79.36 & 83.01 \textcolor{up4}{($\uparrow$4.60\%)} \\
PPE-MMLU & 70.20 & 74.96 \textcolor{up4}{($\uparrow$6.79\%)} & 72.42 & 75.08 \textcolor{up4}{($\uparrow$3.67\%)} & 79.04 & 79.30 \textcolor{up4}{($\uparrow$0.32\%)} \\
RM-B-Code & 60.75 & 69.30 \textcolor{up4}{($\uparrow$14.08\%)} & 67.32 & 70.18 \textcolor{up4}{($\uparrow$4.23\%)} & 72.15 & 73.46 \textcolor{up4}{($\uparrow$1.82\%)} \\
RM-B-Math & 72.78 & 77.69 \textcolor{up4}{($\uparrow$6.75\%)} & 73.72 & 77.79 \textcolor{up4}{($\uparrow$5.51\%)} & 87.81 & 86.20 \textcolor{down4}{($\downarrow$1.83\%)} \\
RB-Reason & 96.65 & 98.25 \textcolor{up4}{($\uparrow$1.66\%)} & 96.30 & 97.48 \textcolor{up4}{($\uparrow$1.23\%)} & 97.52 & 98.15 \textcolor{up4}{($\uparrow$0.64\%)} \\
\midrule
\multicolumn{7}{c}{\textbf{Adversarial Evaluation}} \\
\midrule
RB-Chat-Hard & 84.87 & 87.28 \textcolor{up4}{($\uparrow$2.84\%)} & 84.54 & 83.00 \textcolor{down4}{($\downarrow$1.82\%)} & 84.21 & 84.54 \textcolor{up4}{($\uparrow$0.39\%)} \\
RM-B-Chat & 81.01 & 82.17 \textcolor{up4}{($\uparrow$1.44\%)} & 80.23 & 79.46 \textcolor{down4}{($\downarrow$0.97\%)} & 85.66 & 84.50 \textcolor{down4}{($\downarrow$1.36\%)} \\
OffsetBias & 83.14 & 89.76 \textcolor{up4}{($\uparrow$7.96\%)} & 87.29 & 88.05 \textcolor{up4}{($\uparrow$0.88\%)} & 87.89 & 87.47 \textcolor{down4}{($\downarrow$0.47\%)} \\
\midrule
\multicolumn{7}{c}{\textbf{Safety Evaluation}} \\
\midrule
RMB-Harmless & 69.42 & 70.34 \textcolor{up4}{($\uparrow$1.33\%)} & 69.57 & 68.43 \textcolor{down4}{($\downarrow$1.64\%)} & 60.47 & 59.96 \textcolor{down4}{($\downarrow$0.84\%)} \\
RB-Safety & 92.03 & 93.51 \textcolor{up4}{($\uparrow$1.62\%)} & 93.45 & 93.51 \textcolor{up4}{($\uparrow$0.07\%)} & 92.09 & 92.03 \textcolor{down4}{($\downarrow$0.07\%)} \\
RM-B-Safety-R & 95.54 & 97.13 \textcolor{up4}{($\uparrow$1.67\%)} & 98.41 & 98.41 (0.00\%) & 98.41 & 97.13 \textcolor{down4}{($\downarrow$1.29\%)} \\
RM-B-Safety-F & 99.65 & 99.12 \textcolor{down4}{($\downarrow$0.53\%)} & 98.59 & 98.24 \textcolor{down4}{($\downarrow$0.36\%)} & 96.83 & 97.18 \textcolor{up4}{($\uparrow$0.36\%)} \\
\midrule
\multicolumn{7}{c}{\textbf{Alignment Evaluation}} \\
\midrule
Arena Hard & 88.52 & 89.72 \textcolor{up4}{($\uparrow$1.36\%)} & 89.16 & 90.37 \textcolor{up4}{($\uparrow$1.36\%)} & 90.41 & 90.82 \textcolor{up4}{($\uparrow$0.45\%)} \\
Alpaca Eval & 60.34 & 63.39 \textcolor{up4}{($\uparrow$5.07\%)} & 59.95 & 63.73 \textcolor{up4}{($\uparrow$6.31\%)} & 65.51 & 66.26 \textcolor{up4}{($\uparrow$1.14\%)} \\
Arena Hard-SC & 88.47 & 90.54 \textcolor{up4}{($\uparrow$2.34\%)} & 88.82 & 90.74 \textcolor{up4}{($\uparrow$2.16\%)} & 91.06 & 91.76 \textcolor{up4}{($\uparrow$0.77\%)} \\
Alpaca Eval-LC & 53.26 & 55.45 \textcolor{up4}{($\uparrow$4.12\%)} & 56.83 & 59.80 \textcolor{up4}{($\uparrow$5.23\%)} & 56.95 & 55.58 \textcolor{down4}{($\downarrow$2.40\%)} \\
\bottomrule
\end{tabular}}
\caption{Effect of WorldPM initialization on 72B model's PM fine-tuning performance across different evaluation categories. Abbreviations: RB - RewardBench, RM-B - RM-Bench, Safety-R/F - Safety-Response/Refuse.}
\label{tab:finetune-72b}
\end{table}

\subsection{Evaluation Results}

The results for both the 7B and 72B models are shown in Table~\ref{tab:fintune-7b} and \ref{tab:finetune-72b}, with style-controlled scores reported for all RM benchmarks (see Appendix~\ref{sec:finetuning_without_style} for uncontrolled results). We observe that \textbf{as an initialization for PM fine-tuning, WorldPM universally enhances performance across diverse domains and fine-tuning datasets of varying scales.} The detailed findings are as follows:

\begin{itemize}
\item In subjective domains, PPE-Human and RMB-Helpful show notable improvements, further indicating that WorldPM learns useful general representations for subjective evaluation. RewardBench-Chat shows a slight decrease; however, with accuracy consistently above 97\%, this metric appears saturated and offers limited insight.

\item In objective domains, other PPE metrics, RM-Bench's code and math metrics, and RewardBench's reasoning metrics demonstrate broad improvements. The 72B model shows larger gains compared to 7B, consistent with our WorldPM phase findings: while 7B struggles with objective generalization, 72B exhibits continuous improvement across all objective domains. 

\item In adversarial domains, including Reward bench chat-hard, RM-Bench Chat, and OffsetBias, datasets with initially lower performance (e.g., HelpSteer2) show notable improvements, while others show minimal changes or slight decreases. However, WorldPM itself already achieves high accuracy on these metrics (around 90\% for RM-Bench Chat and OffsetBias, as shown in Figure~\ref{fig:all_style_acc}). The post-fine-tuning decrease might be attributed to simple features being more susceptible to disruption, with final performance approaching that of the fine-tuning datasets.

\item In safety evaluations, including RMB-Harmless, RewardBench-Safety, and RM-Bench-Safety, the 72B model shows mixed performance trends without consistent improvement. This might be related to WorldPM's defensive behavior against pseudo-harmful content in safety domains. Detailed analysis can be found in Appendix~\ref{sec:scaling_tends_all_set}.

\item For downstream alignment evaluation: Comparing Arena Hard and Alpaca Eval results with and without style control, most metrics show improvement. Exceptions occur in RLHFlow dataset's Alpaca Eval, where 7B shows decreased performance in raw results and 72B shows decreased performance after style control. This may be attributed to RLHFlow's large scale and high quality, making WorldPM's benefits less pronounced. 
\end{itemize}

Overall, human preference datasets of different scales all benefit from WorldPM initialization, demonstrating its effectiveness as a starting point. Considering the difficult and expensive annotation process for human preference datasets, large-scale WorldPM serves as a crucial preliminary step before training on these datasets.

\subsection{The Impact of WorldPM Training Scale on Fine-tuning}

To investigate how WorldPM training scale affects fine-tuning benefits, we conduct experiments using different WorldPM checkpoints of the 72B model (trained with 5M, 10M and 15M samples) and the baseline without WorldPM. Each checkpoint served as an initialization point for fine-tuning on Helpsteer2 and Ultrafeedback datasets, with final models selected based on minimum validation loss. Our evaluation encompassed style-controlled PPE metrics (averaging all objective metrics) and downstream alignment performance assessed through Best of 64 sampling on Qwen2.5-72B-Instruct using AlpacaEval and Arena-Hard benchmarks.

\begin{figure}[!htbp]
    \centering
    \includegraphics[width=1.0\linewidth]{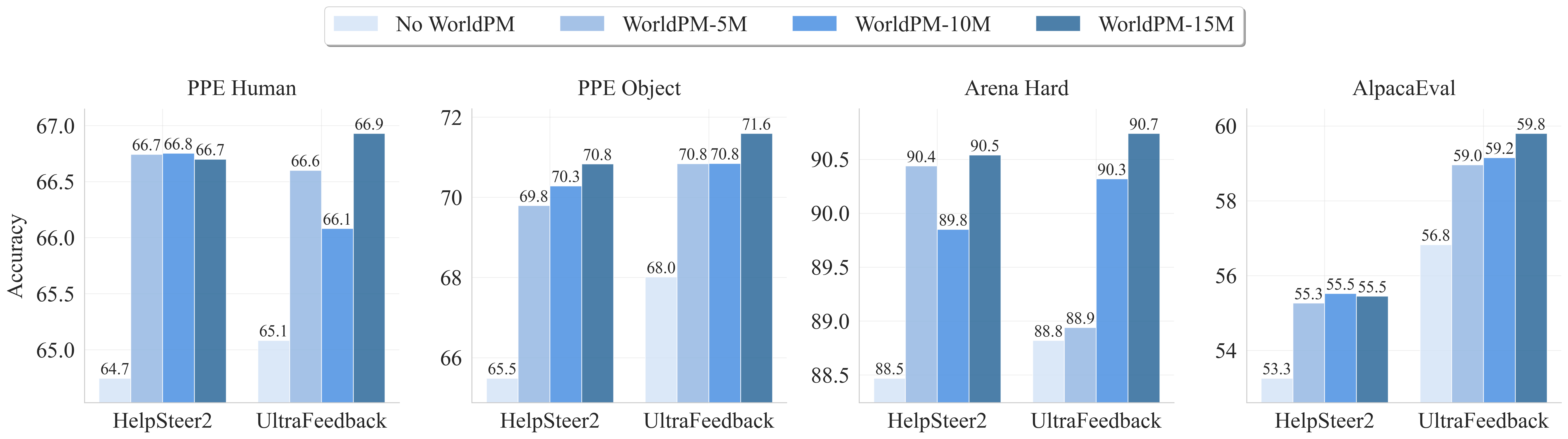}
    \caption{Comparison of PM fine-tuning performance across different WorldPM training scales and baseline without WorldPM. Larger WorldPM scales demonstrate enhanced fine-tuning benefits.}
    \label{fig:scaling_for_pf_mi}
\end{figure}

As illustrated in Figure~\ref{fig:scaling_for_pf_mi}, substantial gains emerge at the 5-million sample threshold, consistent with scaling laws that predict linear performance improvements require exponential growth in training data. Notably, PPE objective metrics exhibit the most consistent improvement with increasing data scale, aligning with our WorldPM phase observations. Other metrics, all of which are subjective evaluations, demonstrate substantial fluctuations; nonetheless, we can observe that \textbf{larger-scale WorldPM consistently achieves better performance on fine-tuning.}

\subsection{Applying WorldPM to RLHF}
We integrate WorldPM into our internal RLHF pipeline, employing GRPO as the RL optimization algorithm. We conduct preference training on two 72B models using in-house preference data, one initialized with WorldPM and the other without. The results are presented in Table~\ref{tab:grpo_compare}. The evaluation includes both in-house benchmarks (first five columns, which are established following the Arena Hard methodology across multiple domains) and open-sourced benchmarks (last three columns). Our findings indicate that WorldPM initialization leads to better alignment with human preferences and improves overall alignment effectiveness.

\begin{table}[htbp]
\small
\centering
\begin{tabular}{lcccccccc}
\toprule

& \multicolumn{5}{c}{\textbf{In-house benchmark}} & \multicolumn{3}{c}{\textbf{Open-sourced benchmark}} \\
\cmidrule(lr){2-6} \cmidrule(lr){7-9}
& Math & Code & OpenQA & Writing & IF & Arena Hard & MT-Bench & Alpaca Eval \\
\midrule
w/o WorldPM & 60.06 & 56.63 & 70.32 & 73.52 & 61.19 & 91.06 & 8.56 & 90.90 \\
w/ WorldPM & \textbf{62.37} & \textbf{64.51} & \textbf{71.42} & \textbf{78.07} & \textbf{66.22} & \textbf{93.13} & \textbf{8.62} & \textbf{91.04} \\
\bottomrule
\end{tabular}
\caption{GRPO alignment results on QwQ-32B: WorldPM-initialized preference models show improved performance across various in-house (Math, Code, OpenQA, Writing, IF) and open-sourced benchmarks. Here, Writing stands for Creative Writing, and IF stands for Instruct Following.}
\label{tab:grpo_compare}
\end{table}

\section{Dicussion}

In this work, our primary contribution extends the exploration in PMP by shifting the focus from evaluating the benefits of reward model (RM) pre-training for downstream RLHF tasks to a deeper investigation of the scalability properties inherent to RM pre-training itself.

Our findings reveal clear scalability trends across objective and adversarial domains, with both parameter count and data volume contributing to performance improvements. 
However, in subjective domains where such scalable trends are not observed, we identify style preference as a potential limiting factor. We find that WorldPM naturally mitigates style preference during training; however, unavoidable style preference within subjective human evaluations can still lead to inappropriate assessments of subjective performance.

This brings us to a fundamental question: what role do RMs play in current systems? Over the past four years, since the introduction of PMP, the role of Reward Models (RMs) has been progressively integrated into more comprehensive reward systems. In objective evaluations, areas such as math and coding now benefit from highly accurate rule-based rewards; factuality assessments increasingly utilize retrieval-augmented methods; and tasks with available references can improve reward signal quality through generative matching techniques. Thanks to extensive engineering efforts, most objective dimensions we care about can be reliably supervised while significantly reducing dependence on RMs.

Thus, while we can continue scaling RM pre-training, how RMs should be integrated with other sources of reward signals remains an open question requiring further exploration. Importantly, in subjective areas—where it is inherently difficult to define accurate rules—the RM's role remains indispensable. However, improving RM performance in subjective domains is less about scaling pre-training data, but rather more about developing better annotation strategies and preference modeling frameworks that go beyond merely capturing surface-level preferences. 
In characterizing human preferences, it is crucial to minimize the subjective cognition of a small number of humans, which is essential for the scalability of preference modeling, as we don't truly understand human preferences ourselves.
Neural networks can only truly understand human preferences when we stop teaching them through conscious labeling and instead incentivize natural alignment with human choices.

\section{Limitaiton}
Our dataset comprises 15M preference pairs (approximately 30G tokens) from StackExchange, modest in contrast to conventional next-token prediction pre-training, which routinely utilizes datasets of trillion-token scale. Significant opportunities exist for expanding preference pretraining data through untapped sources such as various forums and social media platforms.

In addressing subjective evaluation biases, we control only for length and Markdown formatting. Many other influential factors, including emotional preferences and cultural tendencies, remain challenging to capture, underscoring the complexity of subjective assessment. The interconnected nature of various aspects in subjective evaluation makes comprehensive and granular assessment particularly challenging, as these elements prove difficult to isolate effectively.

\bibliographystyle{plainnat}
\bibliography{colm2024_conference}

\appendix

\newpage
\section{Detail of Settings}
\label{sec:detail_of_settings}
\subsection{Forum Data Analysis}
\label{sec:forum_data_analysis}
\begin{figure}
    \centering
    \includegraphics[width=1\linewidth]{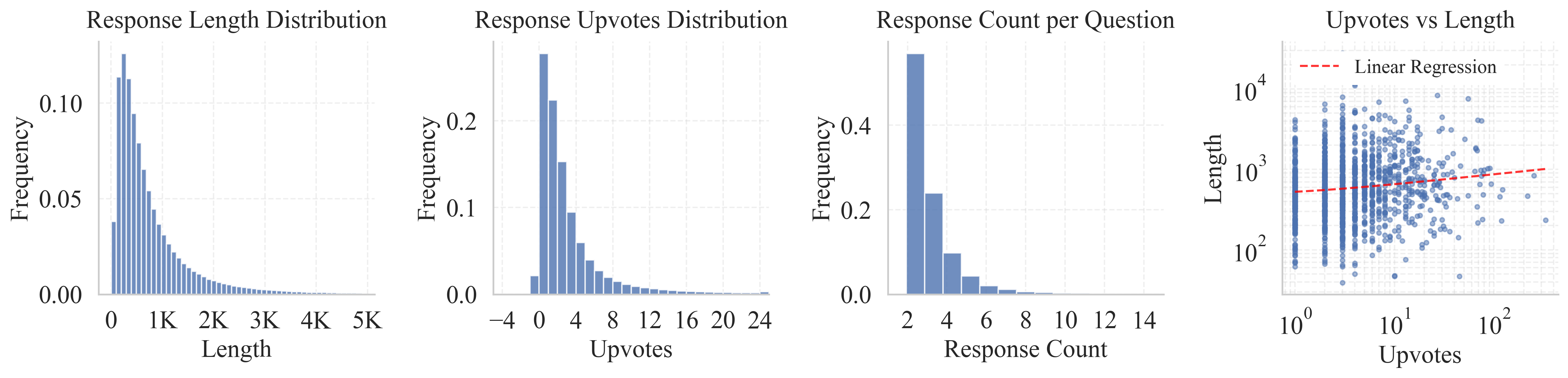}
    \caption{Distribution of reply lengths, upvotes, and the number of replies per post across all StackExchange forum data. All these characteristics demonstrate evident long-tail distributions. The last subplot indicates a weak correlation between reply length and upvotes.}
    \label{fig:se_base_dis}
\end{figure}

\textbf{Different dimensions of forum data generally follow long-tail distributions}. We analyze all the data collected from StackExchange, as shown in Figure~\ref{fig:se_base_dis}. Our analysis focuses on three key dimensions: reply length, reply upvotes, and the number of replies per post. We find that forum data, or naturally generated human data, typically exhibits strong long-tail characteristics. For instance, most forum replies are under 1K in length, receive fewer than 10 upvotes, and posts typically have fewer than 6 replies. Notably, the majority of posts have only one reply, and many replies have zero upvotes. Consequently, we need to filter out a significant portion of data where preference relationships cannot be defined, as establishing a preference pair requires at least two replies to a post with different numbers of upvotes.

It's worth noting that there are relatively few extremely short replies in the forum data, as shown in the first graph. We hypothesize this is due to StackExchange's relatively strict content moderation mechanisms, which filter out replies lacking substantial information.

We further analyze the relationship between reply length and upvotes, as shown in the last column of the figure. Overall, the distribution of reply lengths appears relatively uniform across different upvote counts, although there remains a positive correlation between reply length and upvotes. This correlation is natural in human-generated content, as longer replies typically indicate greater effort from the author, potentially resulting in higher quality content.

\subsection{Preference Data Construction}
\label{sec:preference_data_construction}
On StackExchange, users can upvote or downvote replies, and post authors can choose to accept specific replies. We incorporate the author's acceptance as an additional upvote. The final score for each reply is calculated by subtracting downvotes from upvotes. Using these scores, we define preference directions: when two replies have different scores, we consider them to form a preference relationship.

\textbf{Score differences do not affect preference performance.} One potential concern is whether replies with close scores (e.g., one reply with 1 upvote versus another with 2 upvotes) lack sufficient distinction. To address this, we bucket the score differences between reply pairs into groups: 1-2, 3-5, 6-10, and 11+ differences. As shown in Table~\ref{tab:score_diff_group_compare}, we find similar performance across different benchmarks for each bucket, with no bucket showing significant performance variations.

\begin{table}[!htbp]
\centering
\resizebox{0.8\textwidth}{!}{
\begin{tabular}{lccccc}
\toprule
\textbf{Score Diff}& \textbf{PPE-Human} & \textbf{PPE-Objective} & \textbf{RMB} & \textbf{RewardBench} & \textbf{RM-Bench} \\
\midrule
1-2 & \textbf{63.1} &59.4 &73.8 &79.9 &\textbf{75.2} \\
3-5 & 62.6 &59.5 & \textbf{76.7} & 84.2 & 73.6 \\
6-10 & 62.3 & 59.3 & 76.1 & \textbf{85.5} & 73.2 \\
$\geq$ 11 & 62.5 & \textbf{60.5} & 74.2 & 85.2 & 71.1 \\
\bottomrule
\end{tabular}}
\caption{Bucketing preference pairs by score differences to evaluate the impact of score gaps on preference data quality.}
\label{tab:score_diff_group_compare}
\end{table}

\textbf{Similar performance across different topic domains.} Another potential concern is that different boards of StackExchange, covering diverse content areas, might affect downstream generalization, especially given that our downstream evaluation spans general dialogue, mathematics, coding, and other aspects. To address this, we divide StackExchange data into three segments: StackOverflow (computer science-related topics), Math StackExchange (mathematics-related topics), and Others. This division reflects that the majority of StackExchange content comes from the first two sections, while the remaining 170+ sections contain diverse topics with relatively few questions each.

We train models separately on each section and evaluate them across all test sets, as shown in Table~\ref{tab:board_compare}. Our results indicate no significant performance variations across different sections. Furthermore, StackExchange provides topic tags for each question, allowing for more granular topic identification. For instance, StackOverflow uses specific tags like c++, java, and pointers. These tags also follow a long-tail distribution, with common programming questions constituting the majority. We conduct additional experiments with topic resampling (StackOverflow-TR) on StackOverflow data, deliberately oversampling from less common tags to enhance data diversity. As shown in the table, this resampling approach yields similar results.

These findings suggest that human preferences are domain-agnostic: despite varying discussion topics, users within the same forum community demonstrate consistent preference patterns.

\begin{table}[!htbp]
\centering
\resizebox{0.8\textwidth}{!}{
\begin{tabular}{lccccc}
\toprule[1.5pt]
\textbf{Category} & \textbf{PPE-Human} & \textbf{PPE-Objective} & \textbf{RMB} & \textbf{RewardBench} & \textbf{RM-Bench} \\
\midrule
StackExchange & 62.8 & 62.1 & 76.7 & 84.4 & 72.5 \\
Other & 62.7 & 62.0 & 76.2 & 85.7 & 72.3 \\
Math StackExchange & 62.9 & 62.0 & 75.0 & 83.3 & 75.0 \\
StackOverflow & 63.3 & 63.0 & 75.0 & 83.3 & 74.9 \\
StackOverflow-TR & 62.4 & 62.8 & 75.3 & 84.8 & 73.6 \\
\bottomrule[1.5pt]
\end{tabular}}
\caption{Analysis of performance across different forum categories and topic distributions within individual categories indicates that human preferences exhibit robust cross-domain transferability.}
\label{tab:board_compare}
\end{table}

\subsection{Experimental Settings}
\label{sec:args_settings}
We conduct world preference modeling experiments on Qwen2.5 models ranging from 1.5B to 72B parameters, with a batch size of 10K, training steps of 1536, and a learning rate of 3e-6. For smaller datasets and comparative experiments, we typically set the batch size to 2048. We use the Adam optimizer and employ learning rate warmup with a ratio of 0.1 and the weight decay coefficient of 0.1. We employ learning rate warmup with a ratio of 0.1, followed by a constant learning rate. The context length is set to 2048.

\begin{figure}
    \centering
    \includegraphics[width=1\linewidth]{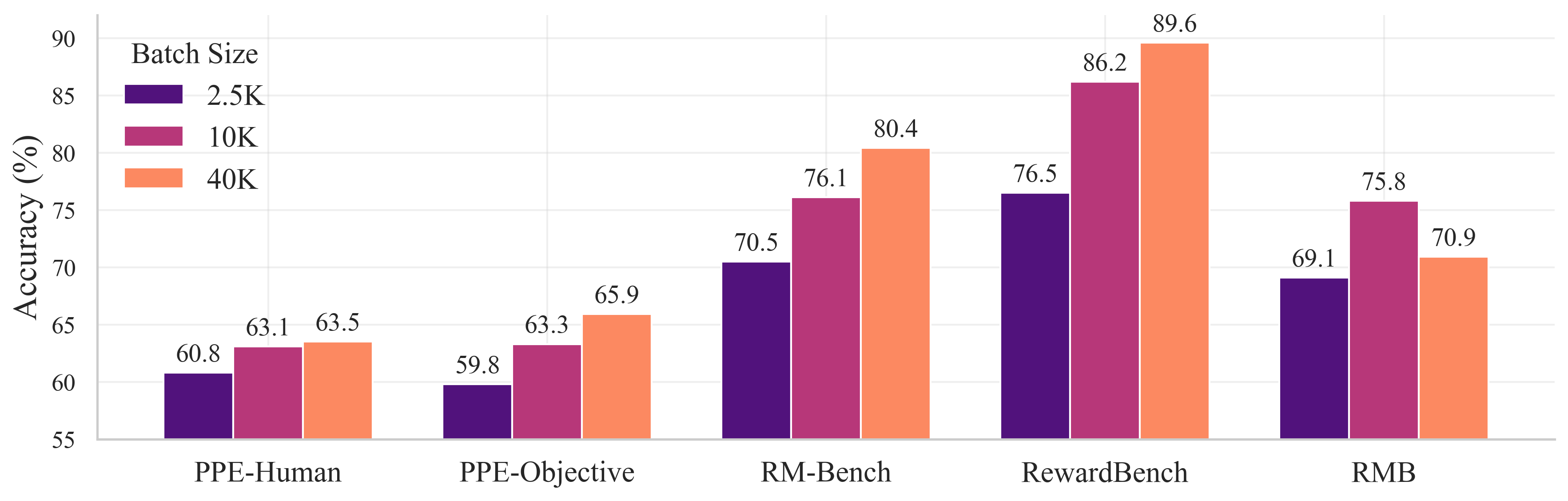}
    \caption{Comparison of different batch sizes (from 2.5K to 40K) on Qwen2.5 7B with fixed 256 steps. Larger batch sizes consistently yield better performance.}
    \label{fig:batch_size_compare}
\end{figure}

We discover that when modeling preferences, larger batch sizes lead to better performance under the same number of training steps, as shown in Figure~\ref{fig:batch_size_compare}. Performance continues to improve even with batch sizes up to 40K, indicating that preference modeling is indeed a challenging task that benefits from more accurate loss estimation through larger batch sizes. However, considering the limited training data available, we ultimately adopt a batch size of 10K to minimize loss noise while maintaining computational efficiency.

\begin{figure}
    \centering
    \includegraphics[width=1\linewidth]{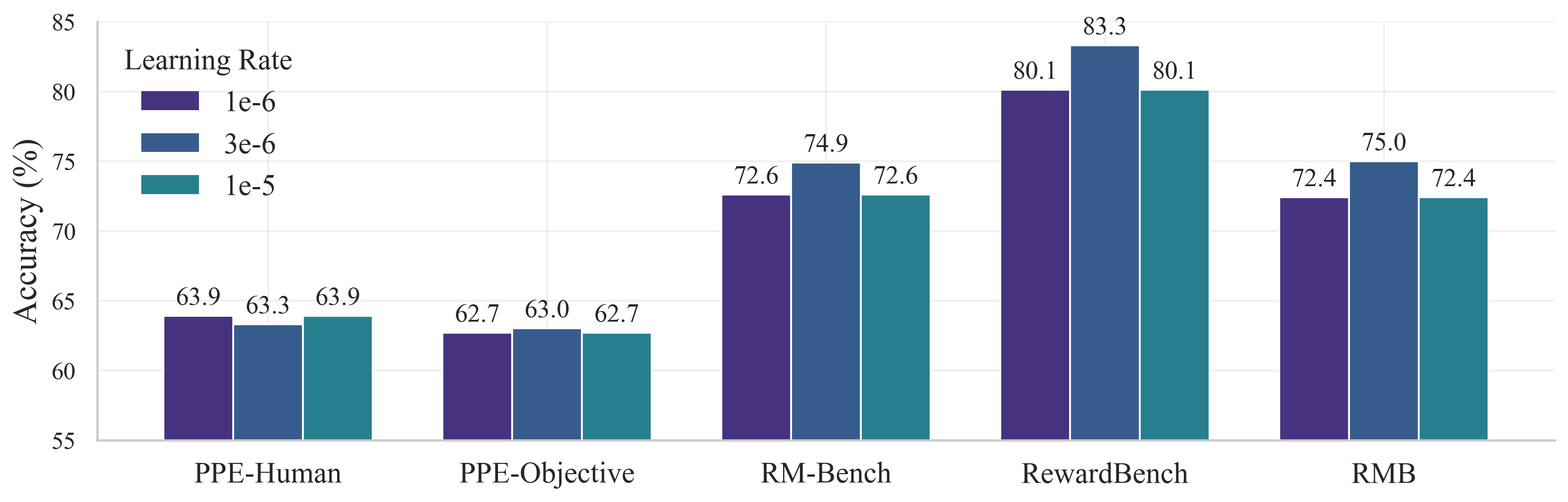}
    \caption{Through our experiments with different learning rates on Qwen2.5 7B, we observe that the model's final performance remains relatively stable across various learning rates, with 3e-6 showing slightly superior results. This suggests that the model's performance is robust to learning rate variations within a reasonable range.}
    \label{fig:lr_compare}
\end{figure}

We additionally conduct comprehensive ablation studies on learning rates, as shown in Figure~\ref{fig:lr_compare}. Our systematic comparison of learning rates at 1e-6, 3e-6, and 1e-5 demonstrates that model performance remains stable across this range, with 3e-6 exhibiting marginally superior results. This finding suggests that preference modeling is robust to learning rate variations within reasonable bounds.

\subsection{Evaluation Settings}
\label{sec:evaluation_settings}
We evaluate our preference models using multiple RM benchmarks, including PPE, RMB, RM-Bench, and RewardBench. However, we modify some of their evaluation methodologies for our specific context.

RMB offers PairWise and Best-of-N (BoN) evaluations, where BoN measures a model's ability to select the best reply from multiple options, and PairWise is essentially BoN with N=2. We focus on PairWise evaluation because: (1) BT loss computation inherently involves pairs of replies, making BoN evaluation computationally challenging, and (2) the original paper shows strong correlation between PairWise and BoN results.

For RM-Bench, we utilize only the model's original responses to form preference pairs, rather than their style-modified versions, as we propose our own method for style-content separation evaluation in this paper.

For PPE, we identify a significant bias issue in the MBPP-Plus test set. This test set is constructed by sampling from four advanced models, where each model generates both correct and incorrect answers to form preference pairs. In our analysis, we observe unusually volatile metrics and conduct a detailed investigation. We define two binary variables: one indicating which answer is correct, and another indicating which answer is longer. 
\begin{wraptable}{R}{0.4\textwidth}
\centering
\begin{tabular}{lc}
\toprule
\textbf{Model} & \textbf{Correlation} \\
\midrule
Gemma-2-9B-IT& $+0.047$ \\
GPT-4o-Mini& $-0.133$ \\
Llama-3-8B-Instruct & $-0.115$ \\
Claude-3-Haiku & $-0.619$ \\
\bottomrule
\end{tabular}
\caption{Correlation coefficients between response length and answer quality across different models.}
\label{tab:length-corr}
\end{wraptable}
By measuring the correlation (Phi coefficient) between answer correctness and length, we find that Claude-3-Haiku samples show a strong negative correlation (-0.6) between these variables, where shorter answers are predominantly correct. This correlation significantly deviates from other models' patterns, indicating a severe style bias. Consequently, we exclude Claude-3-Haiku samples from MBPP-Plus test set to ensure reliable evaluation results.

\subsection{Evaluator Prompts}
The evaluation prompts used in Arena Hard and Alpaca Eval are shown in Figure~\ref{fig:evaluator_prompt_arena_hard} and Figure~\ref{fig:evaluator_prompt_alpaca_eval}, respectively.
\begin{figure}[htbp]
\centering
\begin{tcolorbox}[
    title=Arena Hard's Prompt,
    colback=blue!5,    
    colframe=blue!40,  
]
\begin{lstlisting}
Please act as an impartial judge and evaluate the quality of the responses provided by two AI assistants to the user prompt displayed below. You will be given assistant A's answer and assistant B's answer. Your job is to evaluate which assistant's answer is better.

Begin your evaluation by generating your own answer to the prompt. You must provide your answers before judging any answers.

When evaluating the assistants' answers, compare both assistants' answers with your answer. You must identify and correct any mistakes or inaccurate information.

Then consider if the assistant's answers are helpful, relevant, and concise. Helpful means the answer correctly responds to the prompt or follows the instructions. Note when user prompt has any ambiguity or more than one interpretation, it is more helpful and appropriate to ask for clarifications or more information from the user than providing an answer based on assumptions. Relevant means all parts of the response closely connect or are appropriate to what is being asked. Concise means the response is clear and not verbose or excessive.

Then consider the creativity and novelty of the assistant's answers when needed. Finally, identify any missing important information in the assistants' answers that would be beneficial to include when responding to the user prompt.

After providing your explanation, you must output only one of the following choices as your final verdict with a label:

1. Assistant A is significantly better: [[A>>B]]
2. Assistant A is slightly better: [[A>B]]
3. Tie, relatively the same: [[A=B]]
4. Assistant B is slightly better: [[B>A]]
5. Assistant B is significantly better: [[B>>A]]

Example output: "My final verdict is tie: [[A=B]]".
\end{lstlisting}
\end{tcolorbox}

\caption{Evaluation prompts used in Arena Hard}
\label{fig:evaluator_prompt_arena_hard}
\end{figure}

\begin{figure}[htbp]
\centering
\begin{tcolorbox}[
    title=Alpaca Eval's Prompt,
    colback=blue!5,    
    colframe=blue!40,  
]
\begin{lstlisting}
<|im_start|>system
You are a highly efficient assistant, who evaluates and selects the best large language model (LLMs) based on the quality of their responses to a given instruction. This process will be used to create a leaderboard reflecting the most accurate and human-preferred answers.
<|im_end|>
<|im_start|>user
I require a leaderboard for various large language models. I'll provide you with prompts given to these models and their corresponding outputs. Your task is to assess these responses, and select the model that produces the best output from a human perspective.

## Instruction

{
    "instruction": """{instruction}""",
}
\end{lstlisting}
\begin{lstlisting}
## Model Outputs

Here are the unordered outputs from the models. Each output is associated with a specific model, identified by a unique model identifier.

{
    {
        "model_identifier": "m",
        "output": """{output_1}"""
    },
    {
        "model_identifier": "M",
        "output": """{output_2}"""
    }
}

## Task

Evaluate the models based on the quality and relevance of their outputs, and select the model that generated the best output. Answer by first providing a concise explanation and then end your answer by providing the model identifier of the best output. We will use the last character of your output `output[-1]` as the name of the best model, so make sure you finish with the token of the model identifiers and nothing else: `m` or `M` (no quotes, no dots, no backticks, no new lines, ...). For example:

### Concise explanation
...some text...

### Which is best, m or M?
M

Now is your turn.

## Your answer: "Concise explanation" followed by "Which is best, m or M?"
<|im_end|>

\end{lstlisting}
\end{tcolorbox}

\caption{Evaluation prompts used in Alpaca Eval.}
\label{fig:evaluator_prompt_alpaca_eval}
\end{figure}

\section{Additional Experimental Results}

\subsection{Scaling Trends Across All Test Sets}
\label{sec:scaling_tends_all_set}
\begin{figure}
    \centering
    \includegraphics[width=\linewidth]{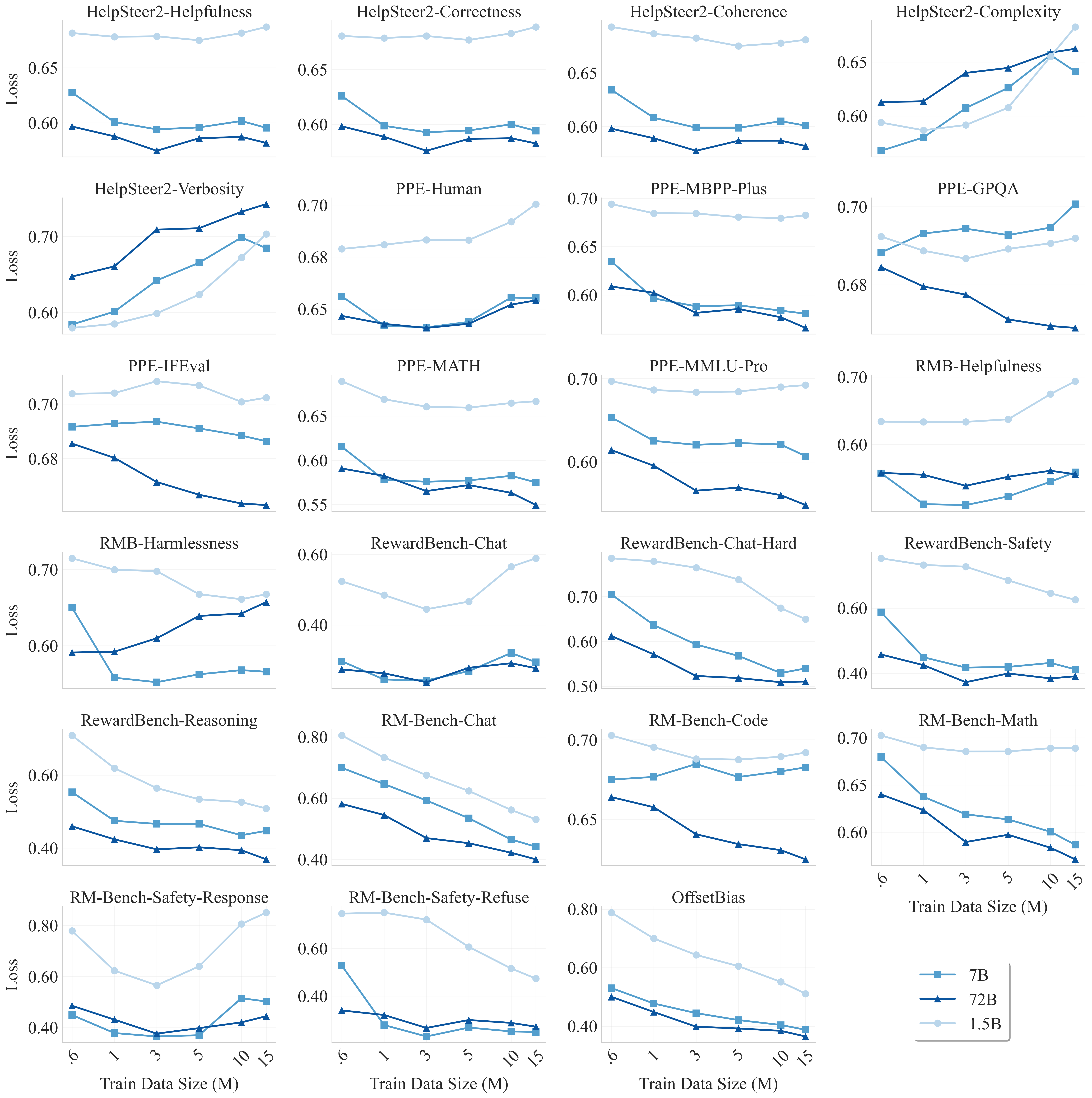}
    \caption{Loss trajectories across evaluation benchmarks as a function of training scale and model size, plotted on $log_2$ scale.}
    \label{fig:all_valid_loss}
\end{figure}

Figure~\ref{fig:all_valid_loss} illustrates the comprehensive scaling trends of world preference across all evaluation benchmarks. In HelpSteer2's five-dimensional assessment framework, we observe strong correlations among Helpfulness, Correctness, and Coherence dimensions, while Complexity and Verbosity demonstrate high mutual correlation. The latter two metrics show strong association with response length, as discussed in Section~\ref{sec:all_test_set_style_control}. As elaborated in the main text, WorldPM demonstrates progressive mitigation of length preference, manifesting as increasing loss on these metrics.

Safety evaluation, omitted from the main text due to space constraints, encompasses RMB-Harmlessness, RewardBench-Safety, and RM-Bench-Safety metrics. RewardBench-Safety exhibits distinct scaling patterns across model sizes: the 1.5B model shows continuous decline, while 7B and 72B models stabilize post-decline, achieving accuracy exceeding 90\%, indicating performance saturation. RMB-Harmlessness presents heterogeneous patterns: declining for 1.5B, saturating for 7B, and ascending for 72B. These diverse patterns can be partially interpreted through RM-Bench-Safety's subdomain analysis: pseudo-harmful queries versus genuinely harmful queries (RM-Bench-Safety-Response), as RMB-Harmlessness also incorporates both aspects. Our analysis reveals decreasing and saturating loss patterns for genuine harmful content detection, while pseudo-harmful query response shows ascending loss in later training stages. This suggests sustained efficacy in identifying genuinely harmful content, coupled with increased conservatism toward pseudo-harmful queries. We hypothesize that seemingly benign queries in specific contexts (e.g., gaming-related weapon discussions) might inadvertently enable harmful applications (e.g., transferable knowledge to real weapons).

\subsection{Preference Fine-tuning Evaluation Without Style Control}
\label{sec:finetuning_without_style}
Tables~\ref{tab:7b-ft-source} and~\ref{tab:72b-ft-source} present the uncontrolled versions of preference fine-tuning results for 7B and 72B models, comparing initialization with and without WorldPM. Overall, the findings align with those observed in the style-controlled versions.

\begin{table}[!htbp]
\centering
\resizebox{0.9\textwidth}{!}{
\begin{tabular}{lcccccc}
\toprule
\multirow{2}{*}{\centering Metrics} & \multicolumn{2}{c}{HelpSteer2} & \multicolumn{2}{c}{UltraFeedback} & \multicolumn{2}{c}{RLHFlow} \\
\cmidrule(lr){2-7}
 & w/o WorldPM & w/ WorldPM & w/o WorldPM & w/ WorldPM & w/o WorldPM & w/ WorldPM \\
\midrule
\multicolumn{7}{c}{\textbf{Subjective Evaluation}} \\
\midrule
PPE-Human & 62.05 & 63.80 \textcolor{up4}{($\uparrow$2.81\%)} & 60.88 & 61.64 \textcolor{up4}{($\uparrow$1.25\%)} & 65.54 & 66.03 \textcolor{up4}{($\uparrow$0.75\%)} \\
RMB-Helpful & 72.49 & 75.84 \textcolor{up4}{($\uparrow$4.62\%)} & 75.62 & 75.70 \textcolor{up4}{($\uparrow$0.11\%)} & 77.64 & 78.33 \textcolor{up4}{($\uparrow$0.88\%)} \\
RB-Chat & 97.49 & 97.21 \textcolor{down4}{($\downarrow$0.29\%)} & 97.49 & 97.21 \textcolor{down4}{($\downarrow$0.29\%)} & 98.04 & 97.77 \textcolor{down4}{($\downarrow$0.28\%)} \\
\midrule
\multicolumn{7}{c}{\textbf{Objective Evaluation}} \\
\midrule
PPE-MBPP & 67.07 & 71.31 \textcolor{up4}{($\uparrow$6.33\%)} & 66.60 & 72.24 \textcolor{up4}{($\uparrow$8.46\%)} & 71.97 & 74.44 \textcolor{up4}{($\uparrow$3.43\%)} \\
PPE-GPQA & 55.41 & 56.56 \textcolor{up4}{($\uparrow$2.08\%)} & 55.62 & 55.18 \textcolor{down4}{($\downarrow$0.81\%)} & 56.76 & 56.70 \textcolor{down4}{($\downarrow$0.10\%)} \\
PPE-IFEval & 57.50 & 56.17 \textcolor{down4}{($\downarrow$2.31\%)} & 56.21 & 56.64 \textcolor{up4}{($\uparrow$0.76\%)} & 61.87 & 62.54 \textcolor{up4}{($\uparrow$1.07\%)} \\
PPE-MATH & 70.80 & 71.74 \textcolor{up4}{($\uparrow$1.32\%)} & 71.64 & 73.77 \textcolor{up4}{($\uparrow$2.97\%)} & 78.79 & 79.94 \textcolor{up4}{($\uparrow$1.46\%)} \\
PPE-MMLU & 63.20 & 66.21 \textcolor{up4}{($\uparrow$4.76\%)} & 65.51 & 66.17 \textcolor{up4}{($\uparrow$1.01\%)} & 67.56 & 70.20 \textcolor{up4}{($\uparrow$3.90\%)} \\
RM-B-Code & 55.92 & 57.68 \textcolor{up4}{($\uparrow$3.14\%)} & 57.02 & 62.06 \textcolor{up4}{($\uparrow$8.85\%)} & 61.40 & 67.76 \textcolor{up4}{($\uparrow$10.36\%)} \\
RM-B-Math & 69.09 & 74.10 \textcolor{up4}{($\uparrow$7.25\%)} & 70.23 & 73.44 \textcolor{up4}{($\uparrow$4.58\%)} & 75.99 & 79.58 \textcolor{up4}{($\uparrow$4.73\%)} \\
RB-Reason & 81.62 & 96.12 \textcolor{up4}{($\uparrow$17.77\%)} & 83.19 & 91.86 \textcolor{up4}{($\uparrow$10.42\%)} & 94.13 & 96.16 \textcolor{up4}{($\uparrow$2.15\%)} \\
\midrule
\multicolumn{7}{c}{\textbf{Adversarial Evaluation}} \\
\midrule
RB-Chat-Hard & 62.50 & 74.89 \textcolor{up4}{($\uparrow$19.82\%)} & 61.51 & 67.32 \textcolor{up4}{($\uparrow$9.45\%)} & 66.01 & 69.96 \textcolor{up4}{($\uparrow$5.98\%)} \\
RM-B-Chat & 51.94 & 65.12 \textcolor{up4}{($\uparrow$25.37\%)} & 57.36 & 66.67 \textcolor{up4}{($\uparrow$16.22\%)} & 56.98 & 66.67 \textcolor{up4}{($\uparrow$17.01\%)} \\
OffsetBias & 63.76 & 82.40 \textcolor{up4}{($\uparrow$29.23\%)} & 72.98 & 78.17 \textcolor{up4}{($\uparrow$7.11\%)} & 73.54 & 80.50 \textcolor{up4}{($\uparrow$9.47\%)} \\
\midrule
\multicolumn{7}{c}{\textbf{Safety Evaluation}} \\
\midrule
RMB-Harmless & 68.41 & 73.58 \textcolor{up4}{($\uparrow$7.56\%)} & 69.22 & 70.23 \textcolor{up4}{($\uparrow$1.46\%)} & 66.78 & 66.84 \textcolor{up4}{($\uparrow$0.08\%)} \\
RB-Safety & 82.84 & 90.54 \textcolor{up4}{($\uparrow$9.30\%)} & 89.05 & 87.84 \textcolor{down4}{($\downarrow$1.37\%)} & 85.95 & 86.35 \textcolor{up4}{($\uparrow$0.47\%)} \\
RM-B-Safety-R & 87.90 & 92.99 \textcolor{up4}{($\uparrow$5.80\%)} & 95.54 & 94.27 \textcolor{down4}{($\downarrow$1.33\%)} & 91.08 & 93.63 \textcolor{up4}{($\uparrow$2.80\%)} \\
RM-B-Safety-F & 98.59 & 97.54 \textcolor{down4}{($\downarrow$1.07\%)} & 96.48 & 96.48 (0.00\%) & 95.77 & 95.42 \textcolor{down4}{($\downarrow$0.37\%)} \\
\bottomrule
\end{tabular}}
\caption{Effect of WorldPM initialization on 7B model's PM fine-tuning performance across different evaluation categories (without style control). Abbreviations: RB - RewardBench, RM-B - RM-Bench, Safety-R/F - Safety-Response/Refuse.}
\label{tab:7b-ft-source}
\end{table}

\begin{table}[!htbp]
\centering
\resizebox{0.9\textwidth}{!}{
\begin{tabular}{lcccccc}
\toprule
\multirow{2}{*}{\centering Metrics} & \multicolumn{2}{c}{HelpSteer2} & \multicolumn{2}{c}{UltraFeedback} & \multicolumn{2}{c}{RLHFlow} \\
\cmidrule(lr){2-7}
 & w/o WorldPM & w/ WorldPM & w/o WorldPM & w/ WorldPM & w/o WorldPM & w/ WorldPM \\
\midrule
\multicolumn{7}{c}{\textbf{Subjective Evaluation}} \\
\midrule
PPE-Human & 62.22 & 65.53 \textcolor{up4}{($\uparrow$5.32\%)} & 62.09 & 64.51 \textcolor{up4}{($\uparrow$3.89\%)} & 67.63 & 68.29 \textcolor{up4}{($\uparrow$0.98\%)} \\
RMB-Helpful & 76.40 & 77.45 \textcolor{up4}{($\uparrow$1.38\%)} & 75.60 & 79.10 \textcolor{up4}{($\uparrow$4.64\%)} & 78.10 & 80.08 \textcolor{up4}{($\uparrow$2.53\%)} \\
RB-Chat & 97.91 & 98.88 \textcolor{up4}{($\uparrow$1.00\%)} & 97.21 & 96.09 \textcolor{down4}{($\downarrow$1.15\%)} & 98.32 & 98.04 \textcolor{down4}{($\downarrow$0.28\%)} \\
\midrule
\multicolumn{7}{c}{\textbf{Objective Evaluation}} \\
\midrule
PPE-MBPP & 70.19 & 77.45 \textcolor{up4}{($\uparrow$10.34\%)} & 74.32 & 78.65 \textcolor{up4}{($\uparrow$5.82\%)} & 81.00 & 81.85 \textcolor{up4}{($\uparrow$1.05\%)} \\
PPE-GPQA & 58.03 & 59.28 \textcolor{up4}{($\uparrow$2.15\%)} & 59.06 & 60.76 \textcolor{up4}{($\uparrow$2.88\%)} & 62.19 & 63.44 \textcolor{up4}{($\uparrow$2.01\%)} \\
PPE-IFEval & 58.96 & 63.50 \textcolor{up4}{($\uparrow$7.68\%)} & 60.16 & 63.83 \textcolor{up4}{($\uparrow$6.10\%)} & 63.32 & 65.04 \textcolor{up4}{($\uparrow$2.71\%)} \\
PPE-MATH & 69.73 & 76.64 \textcolor{up4}{($\uparrow$9.92\%)} & 72.58 & 77.89 \textcolor{up4}{($\uparrow$7.32\%)} & 79.10 & 82.52 \textcolor{up4}{($\uparrow$4.32\%)} \\
PPE-MMLU & 69.98 & 74.69 \textcolor{up4}{($\uparrow$6.73\%)} & 72.07 & 74.92 \textcolor{up4}{($\uparrow$3.96\%)} & 78.52 & 79.28 \textcolor{up4}{($\uparrow$0.97\%)} \\
RM-B-Code & 61.18 & 69.30 \textcolor{up4}{($\uparrow$13.26\%)} & 65.79 & 68.64 \textcolor{up4}{($\uparrow$4.33\%)} & 71.93 & 72.81 \textcolor{up4}{($\uparrow$1.22\%)} \\
RM-B-Math & 70.04 & 76.75 \textcolor{up4}{($\uparrow$9.58\%)} & 71.55 & 77.32 \textcolor{up4}{($\uparrow$8.06\%)} & 87.52 & 85.63 \textcolor{down4}{($\downarrow$2.16\%)} \\
RB-Reason & 96.61 & 98.22 \textcolor{up4}{($\uparrow$1.66\%)} & 95.70 & 96.44 \textcolor{up4}{($\uparrow$0.77\%)} & 96.33 & 97.38 \textcolor{up4}{($\uparrow$1.09\%)} \\
\midrule
\multicolumn{7}{c}{\textbf{Adversarial Evaluation}} \\
\midrule
RB-Chat-Hard & 69.41 & 80.26 \textcolor{up4}{($\uparrow$15.64\%)} & 75.66 & 77.63 \textcolor{up4}{($\uparrow$2.61\%)} & 74.12 & 75.33 \textcolor{up4}{($\uparrow$1.63\%)} \\
RM-B-Chat & 77.52 & 75.97 \textcolor{down4}{($\downarrow$2.00\%)} & 70.93 & 70.54 \textcolor{down4}{($\downarrow$0.55\%)} & 77.13 & 75.58 \textcolor{down4}{($\downarrow$2.01\%)} \\
OffsetBias & 76.29 & 85.79 \textcolor{up4}{($\uparrow$12.46\%)} & 84.47 & 85.86 \textcolor{up4}{($\uparrow$1.64\%)} & 83.18 & 81.90 \textcolor{down4}{($\downarrow$1.54\%)} \\
\midrule
\multicolumn{7}{c}{\textbf{Safety Evaluation}} \\
\midrule
RMB-Harmless & 68.96 & 70.10 \textcolor{up4}{($\uparrow$1.66\%)} & 68.58 & 67.93 \textcolor{down4}{($\downarrow$0.95\%)} & 63.70 & 62.65 \textcolor{down4}{($\downarrow$1.64\%)} \\
RB-Safety & 90.14 & 91.35 \textcolor{up4}{($\uparrow$1.35\%)} & 89.12 & 90.61 \textcolor{up4}{($\uparrow$1.67\%)} & 84.19 & 83.24 \textcolor{down4}{($\downarrow$1.12\%)} \\
RM-B-Safety-R & 94.59 & 94.27 \textcolor{down4}{($\downarrow$0.34\%)} & 97.45 & 98.09 \textcolor{up4}{($\uparrow$0.65\%)} & 94.90 & 93.63 \textcolor{down4}{($\downarrow$1.34\%)} \\
RM-B-Safety-F & 97.89 & 97.18 \textcolor{down4}{($\downarrow$0.72\%)} & 97.01 & 96.48 \textcolor{down4}{($\downarrow$0.54\%)} & 88.73 & 89.08 \textcolor{up4}{($\uparrow$0.40\%)} \\
\bottomrule
\end{tabular}}
\caption{Effect of WorldPM initialization on 72B model's PM fine-tuning performance across different evaluation categories (without style control). Abbreviations: RB - RewardBench, RM-B - RM-Bench, Safety-R/F - Safety-Response/Refuse.}
\label{tab:72b-ft-source}
\end{table}

\section{Details of Style Control}
\label{sec:detail_of_style_control}
\subsection{Ablation Study on Style Control Factors}
\begin{figure}
    \centering
    \includegraphics[width=.8\linewidth]{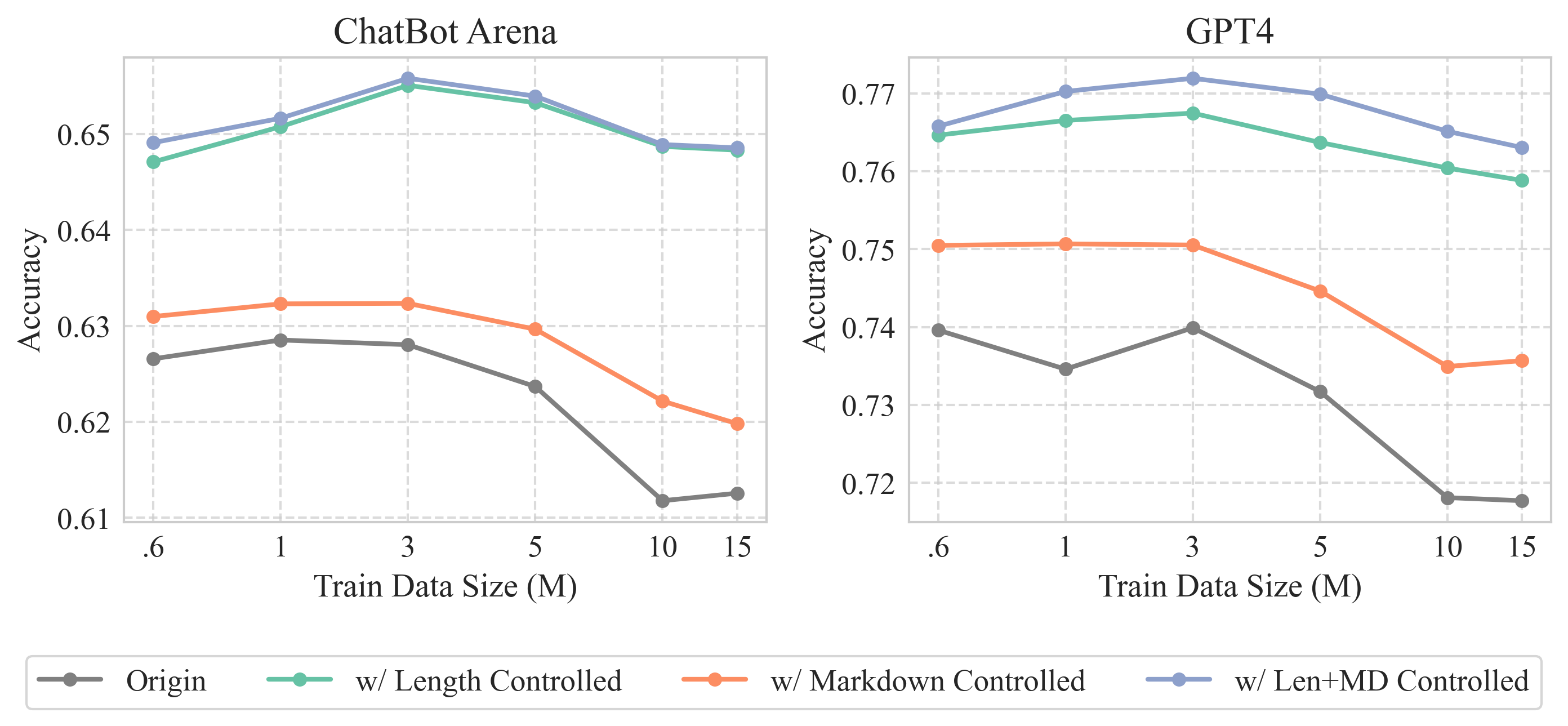}
    \caption{Impact of controlling different style factors on subjective evaluation performance.}
    \label{fig:style_control_compare}
\end{figure}

We control two style features: length and markdown formatting. As shown in Figure~\ref{fig:style_control_compare}, we compare the impact on subjective evaluation performance when controlling length alone, markdown formatting alone, and both factors simultaneously. Our findings indicate that markdown formatting has less influence than length, further confirming that length is the primary style factor. Moreover, modeling both factors together enables more effective style control.

\subsection{Effect of Style Control Across Test Set Evaluation}
\label{sec:all_test_set_style_control}
\begin{figure}
    \centering
    \includegraphics[width=1\linewidth]{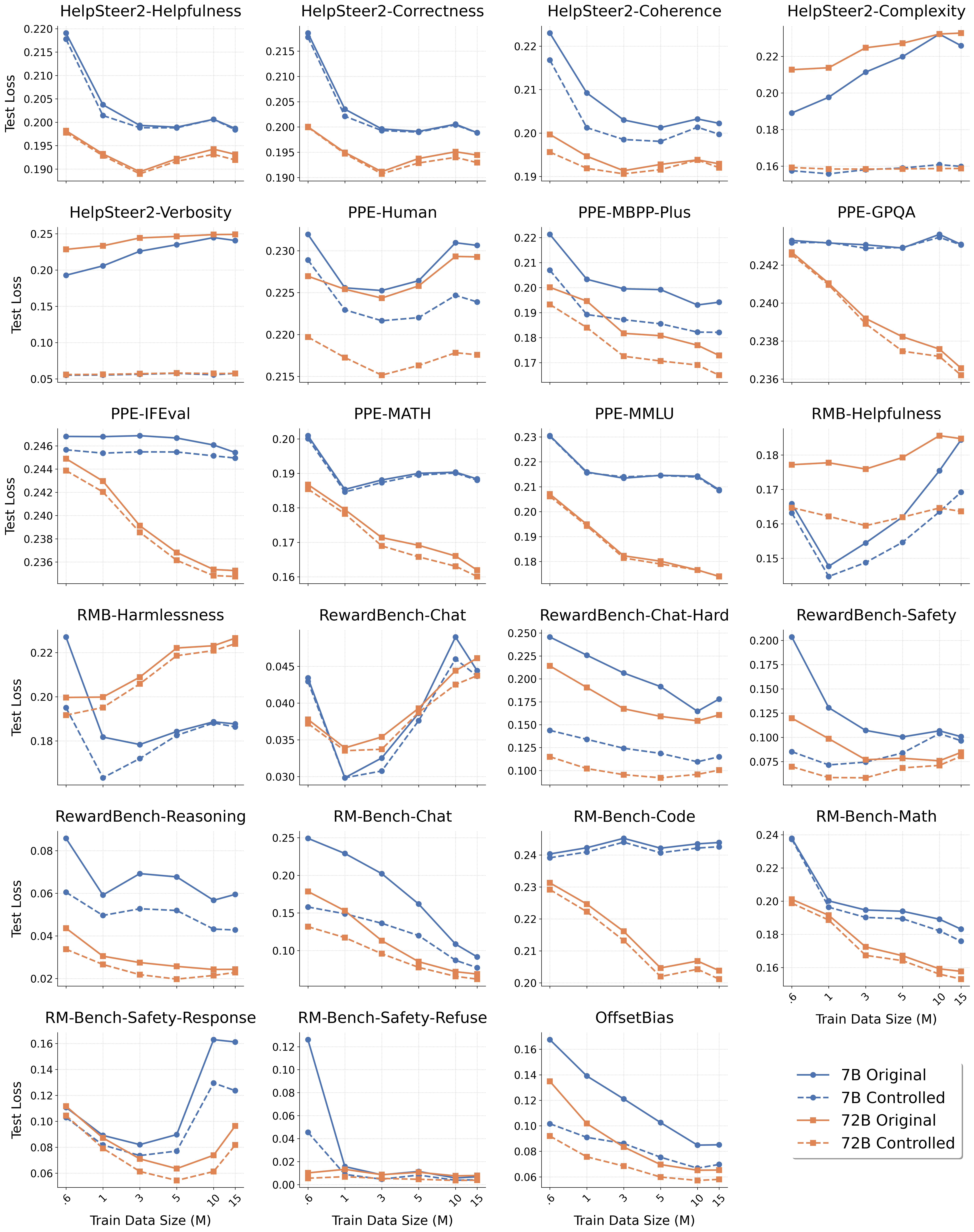}
    \caption{Effect of style control on test loss across all test set evaluation.}
    \label{fig:all_style_loss}
\end{figure}

\begin{figure}
    \centering
    \includegraphics[width=1\linewidth]{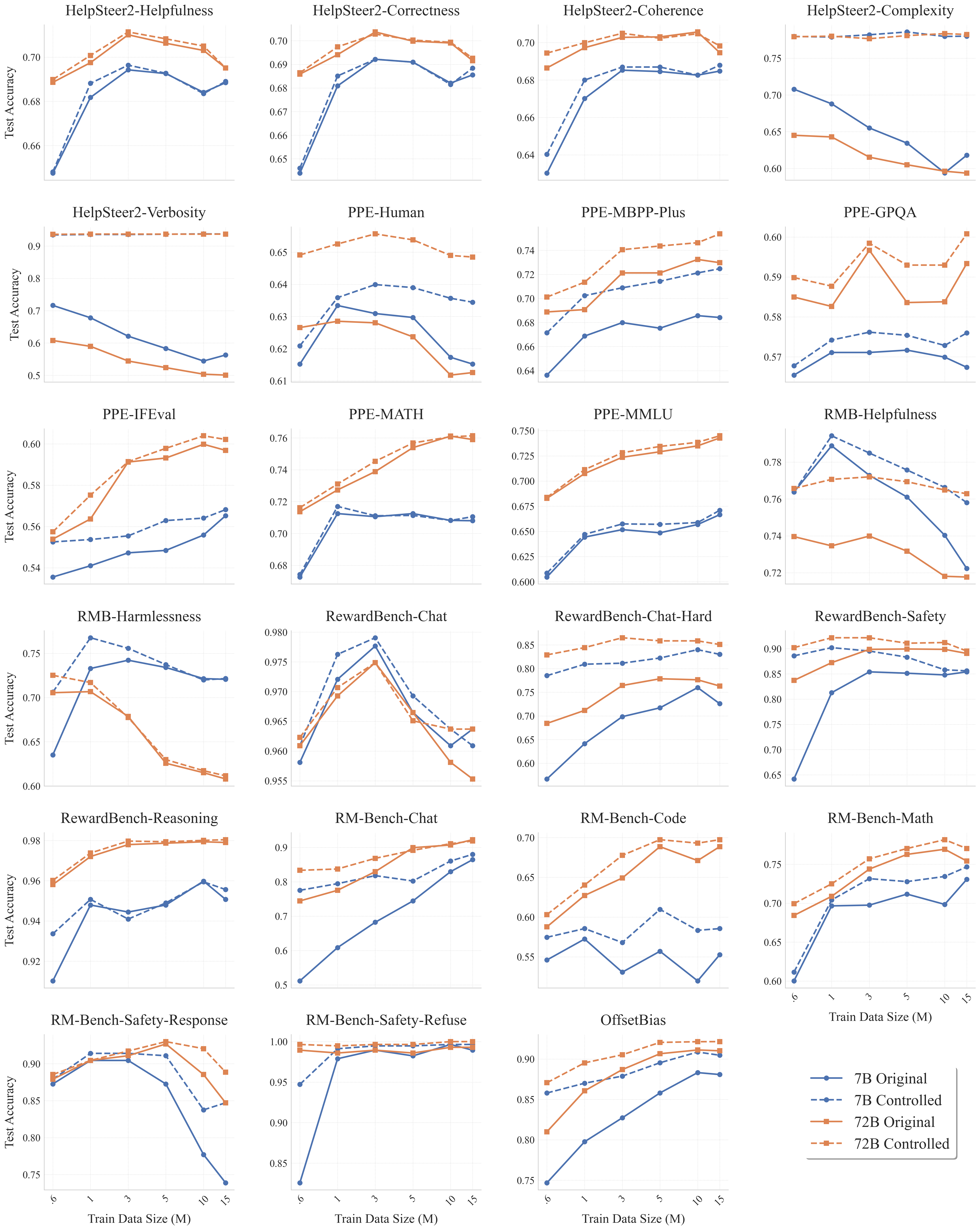}
    \caption{Effect of style control on performance across all test set evaluation.}
    \label{fig:all_style_acc}
\end{figure}

Figure~\ref{fig:all_style_acc} and Figure~\ref{fig:all_style_loss} demonstrates the impact of style control on evaluation results (test accuracy and test loss) across all test sets. We observe several key findings:

\begin{itemize}
\item In HelpSteer2, the dimensions of Helpfulness, Correctness, and Coherence maintain substantial consistency. Complexity and Verbosity exhibit high correlation with length characteristics. Upon implementing length control, these two metrics show significant changes while maintaining stable trends.

\item Some objective and robustness evaluation sets demonstrate notable shifts in performance (e.g., PPE-MBPP-Plus, PPE-GPQA, and RM-Bench-Chat). However, the performance gap between style-controlled and uncontrolled versions either remains constant or narrows with increased training scale, contrasting with the widening gap in subjective domains. This suggests that modeling world preference gradually overcomes length bias, converging toward more accurate evaluation results.

\item In the safety domain, RM-Bench-Safety-Refuse demonstrates exceptionally high accuracy even without style control, indicating WorldPM's inherent capability for safety discrimination. Other safety-related benchmarks show declining accuracy in later stages, potentially related to increased defense against pseudo-harmful queries.
\end{itemize}

\section{Reflections on Potential Noise in Forum Data}

\subsection{Impact of RM Filtering on Different Data Sources}
In the early stages of our research, we hypothesized that forum preference data contained significant noise, leading us to invest considerable effort in denoising attempts. One primary approach involved using existing reward models for noise reduction. Specifically, we applied Qwen-2.5-72B-RM~\citep{Yang2024Qwen25TR} to score preference data from Quora and StackExchange, filtering out instances where the Chosen score was lower than the Rejected score. This approach effectively creates an intersection of two preference sources: human preferences and RM preferences. We conducted training experiments on the 7B model, with results shown in Table~\ref{tab:dataset-comparison}.

Our findings reveal that for Quora data, RM-based filtering significantly improves performance across all benchmarks. However, for StackExchange data, the improvements are less pronounced, with notable gains only in PPE-Objective. This disparity suggests that StackExchange's inherently stricter quality control mechanisms result in more reliable data, making additional filtering less impactful.

\begin{table}[!htbp]
\centering
\begin{tabular}{lccccc}
\toprule
\textbf{Dataset} & \textbf{PPE-Human} & \textbf{PPE-Objective} & \textbf{RMB} & \textbf{RewardBench} & \textbf{RM-Bench} \\
\midrule
Quora & 60.2 & 57.7 & 66.1 & 69.0 & 60.9 \\
Quora+Filter & 62.2 & 62.9 & 76.1 & 85.4 & 74.4 \\
\midrule
StackExchange & 62.8 & 62.1 & 76.7 & 84.4 & 72.5 \\
StackExchange+Filter & 63.5 & 64.1 & 76.3 & 84.9 & 73.5 \\
\bottomrule
\end{tabular}
\caption{Impact of RM-based Filtering on Preference Data from Quora and StackExchange.}
\label{tab:dataset-comparison}
\end{table}

\subsection{Impact of Various Reward Model Filtering}
We expand the StackExchange training dataset to 2 million samples and employ two state-of-the-art models, Athene-70B-RM~\citep{Athene2024} and Qwen-2.5-72B-RM~\citep{Yang2024Qwen25TR}, for both independent and joint filtering before training on the 7B model to further investigate the impact of RM filtering. As shown in Table~\ref{tab:se-model-comparison}, while filtering operations yield modest improvements across different benchmarks, the overall impact remains limited. Notably, even joint filtering using both models fails to demonstrate significant performance gains, despite maintaining a considerable performance gap compared to the filtering models themselves.
\begin{table}[!htbp]
\centering
\begin{tabular}{lccccc}
\toprule
\textbf{Dataset} & \textbf{PPE-Human} & \textbf{PPE-Objective} & \textbf{RMB} & \textbf{RewardBench} & \textbf{RM-Bench} \\
\midrule
StackExchange & 63.1 & 63.2 & 75.6 & 85.9 & 74.0 \\
\midrule
Athene70B & 66.4 & 70.5 & 80.7 & 88.3 & 79.4 \\
Athene70B-Filter & 63.7 & 65.1 & 77.1 & 86.0 & 75.3 \\
\midrule
Qwen72B & 63.7 & 72.8 & 72.3 & 91.9 & 84.5 \\
Qwen72B-Filter & 63.1 & 65.1 & 76.9 & 86.9 & 75.9 \\
\midrule
Athene70B-Qwen72B-Filter & 63.6 & 65.5 & 77.5 & 85.7 & 75.6 \\
\bottomrule
\end{tabular}
\caption{Comparison of Independent and Joint Filtering Effects Using Athene-70B-RM and Qwen-2.5-72B-RM on StackExchange Data, showing limited gains despite performance gap with original models.}
\label{tab:se-model-comparison}
\end{table}

\subsection{Potential Biases in RM Filtering}
We bucket the scoring results from Qwen-2.5-72B-RM by calculating the score difference between Chosen and Rejected responses for each sample to investigate the distinctions between filtered and filtered-out data. As shown in Table~\ref{tab:score-range-comparison}, we categorize score differences into various intervals. We find that filtered data shows relatively consistent performance across different intervals, with only the 0\~0.7 interval performing marginally worse than others.

For filtered-out data (score range -10.0~0), we present their evaluation results alongside their inverse scores (100 minus the original score) in parentheses. Notably, the performance of these filtered-out data closely aligns with that of the filtered data, suggesting that both sets possess similar modeling capabilities but with opposing preference directions.

We conclude that the filtering operation essentially aligns with RM's assessment patterns, retaining data that conforms to its discrimination criteria while filtering out non-conforming instances. However, this implies that RM-based filtering may lead models to adopt RM's own discrimination patterns. Therefore, we argue that applying RM filtering diverges from capturing world preference. \textbf{Instead of assuming forum data contains noise, we should interpret apparent contradictions as manifestations of genuine human preferences, allowing models to discover underlying commonalities within these surface-level conflicts.}

\begin{table}[!htbp]
\centering
\begin{tabular}{lccccc}
\toprule
\textbf{Score Range} & \textbf{PPE-Human} & \textbf{PPE-Objective} & \textbf{RMB} & \textbf{RewardBench} & \textbf{RM-Bench} \\
\midrule
$[2.8, 10.0]$ & 62.8 & 64.6 & 76.6 & 85.4 & 75.2 \\
$[1.6, 2.8]$ & 63.4 & 64.8 & 76.9 & 85.7 & 75.8 \\
$[0.7, 1.6]$ & 63.4 & 64.1 & 76.4 & 84.3 & 74.9 \\
$[0.0, 0.7]$ & 62.9 & 62.4 & 76.2 & 84.5 & 74.0 \\
$[-1.6, 0]$ & 38.1 (61.9) & 36.0 (64.0) & 24.9 (75.1) & 20.3 (79.7) & 23.7 (76.3) \\
$[-10.0, -1.6]$ & 37.3 (62.7) & 35.0 (65.0) & 23.0 (77.0) & 15.8 (84.2)& 24.6 (75.4) \\
\bottomrule
\end{tabular}
\caption{Performance comparison across different RM score ranges and metrics.}
\label{tab:score-range-comparison}
\end{table}

\section{Case Study}
\label{sec:case_study}
We present several examples from PPE-Human, including Figures~\ref{fig:example_ppe_1}, \ref{fig:example_ppe_2}, and~\ref{fig:example_ppe_3}. In Figures~\ref{fig:example_ppe_1} and~\ref{fig:example_ppe_2}, general humans chose incorrect answers, while WorldPM correctly identified the errors. Figures~\ref{fig:example_ppe_3} demonstrates that data from general humans can be low-quality and uninformative, thus interfering with evaluation.

We also present several examples of training data used for modeling world preference, including Figures~\ref{fig:example_se_1}, \ref{fig:example_se_2}, \ref{fig:example_se_3}, and~\ref{fig:example_se_4}. These data often contain rich information, enabling the model to learn complex representations.

\begin{figure}[htbp]
\centering
\begin{tcolorbox}[
    title=Query,
    colback=gray!5,        
    colframe=gray!50,      
    coltitle=black,       
    width=\textwidth
]

Prove or disprove: for all integers $a$, $b$, $c$, $d$, if $a \mid b$ and $c \mid d$, then $(a + c) \mid (b + d)$.
\end{tcolorbox}
\begin{tcolorbox}[
    title=Chosen Response (World Preference Model Score: -1.84),
    colback=red!2,         
    colframe=red!35,       
    coltitle=black        
]
This statement is true. This is because absolute values of sums of integers are equal to the sums of their absolute values. 

In this case, since $a \mid b$ ($a$ divides $b$) and $c \mid d$ ($c$ divides $d$), then certainly $|b|$ and $|d|$ are both divisible by $|a|$ and $|c|$, respectively. So, setting $e = |a|$ and $f = |c|$, we have that both $b = q \cdot e$ for some integer $q$ and $d = r \cdot f$ for some integer $r$. Then, $b+d = (q+r) \cdot e+f$ is divisible by both $e$ and $f$, which are equal to $|a|$ and $|c|$. So $(b+d)$ is divisible by the sum $|a+c|$. Thus, $(a+c) \mid (b+d)$.
\end{tcolorbox}
\begin{tcolorbox}[
    title=Rejected Response (World Preference Model Score: 0.76),
    colback=green!2,       
    colframe=green!35,     
    coltitle=black         
]
To prove or disprove the statement "for all integers $a$, $b$, $c$, $d$ if $a \mid b$ and $c \mid d$, then $a + c \mid b + d$", we can approach it by first understanding what the statement means and then testing it with specific examples.

\textbf{Definitions:}
- $a \mid b$ means that there exists an integer $k$ such that $b = ak$.
- $c \mid d$ means that there exists an integer $m$ such that $d = cm$.

\textbf{Claim:}
- $a + c \mid b + d$ means there exists an integer $n$ such that $b + d = (a + c)n$.

We need to see if from $b = ak$ and $d = cm$, it follows that $b + d = (a + c)n$ for some integer $n$.

\textbf{Counterexample:}
Let's consider specific numbers to see if this statement holds.
- Choose $a = 2$, $b = 4$, $c = 3$, $d = 6$.
- Clearly, $2 \mid 4$ because $4 = 2 \times 2$.
- Clearly, $3 \mid 6$ because $6 = 3 \times 2$.
- Now, $a + c = 2 + 3 = 5$ and $b + d = 4 + 6 = 10$.

We need to check if $5 \mid 10$. Since $10 = 5 \times 2$, in this case, $5$ does actually divide $10$. However, we need to check more examples to see if there are any that fail.

\textbf{Another Example:}
- Choose $a = 2$, $b = 4$, $c = 3$, $d = 9$.
- Again, $2 \mid 4$ since $4 = 2 \times 2$.
- $3 \mid 9$ since $9 = 3 \times 3$.
- Calculate $a + c = 2 + 3 = 5$ and $b + d = 4 + 9 = 13$.

Now, we need to check if $5 \mid 13$. Since $13$ is not divisible by $5$ (there is no integer $n$ such that $13 = 5n$), this proves that $5$ does not divide $13$.

\textbf{Conclusion:}
The counterexample where $a = 2$, $b = 4$, $c = 3$, $d = 9$ disproves the statement. Even though $a$ divides $b$ and $c$ divides $d$, $a + c$ does not necessarily divide $b + d$. Thus, the statement "for all integers $a$, $b$, $c$, $d$ if $a \mid b$ and $c \mid d$, then $a + c \mid b + d$" is false.
\end{tcolorbox}

\caption{Example from the PPE-Human test set where the chosen response's proof is wrong while the rejected response's proof is correct, and the world preference model identify this error.}
\label{fig:example_ppe_1}
\end{figure}

\begin{figure}[htbp]
\centering

\begin{tcolorbox}[
    title=Query,
    colback=gray!5,        
    colframe=gray!50,      
    coltitle=black,        
    width=\textwidth
]
userscript to copy the text of the hovered link when the key combination alt+c is pressed. just present the code and make it concise.
\end{tcolorbox}

\begin{tcolorbox}[
    title=Chosen Response (World Preference Model Score: -0.66),
    colback=red!2,         
    colframe=red!35,      
    coltitle=black         
]
\begin{lstlisting}
// @require     https://code.jquery.com/jquery-3.6.0.min.js

$(document).keydown(function(event){
    if(event.altKey && event.keyCode == 67){
        var link = $("hoveredElement").text();
        clipboard.writeText(link);
    }
});
\end{lstlisting}
\end{tcolorbox}

\begin{tcolorbox}[
    title=Rejected Response (World Preference Model Score: 1.42),
    colback=green!2,      
    colframe=green!35,    
    coltitle=black        
]
\begin{lstlisting}
// ==UserScript==
// @name         Copy Hovered Link Text
// @namespace    http://tampermonkey.net/
// @version      1.0
// @description  Copy the text of the hovered link when Alt+C is pressed
// @author       You
// @match        *://*/*
// @grant        none
// ==/UserScript==

(function() {
    'use strict';
    
    let hoveredLinkText = '';

    document.addEventListener('mouseover', function(e) {
        if (e.target.tagName.toLowerCase() === 'a') {
            hoveredLinkText = e.target.innerText || e.target.textContent;
        }
    });

    document.addEventListener('keydown', function(e) {
        if (e.altKey && e.key === 'c') {
            if (hoveredLinkText) {
                navigator.clipboard.writeText(hoveredLinkText).then(() => {
                    console.log('Text copied to clipboard: ', hoveredLinkText);
                }).catch(err => {
                    console.error('Failed to copy text: ', err);
                });
            }
        }
    });
})();
\end{lstlisting}
\end{tcolorbox}

\caption{Example from the PPE-Human test set, where Chosen uses an invalid selector \$("hoveredElement"), while Rejected provides a correct solution.}
\label{fig:example_ppe_2}
\end{figure}

\begin{figure}[htbp]
\centering

\begin{tcolorbox}[
    title=Query,
    colback=gray!5,        
    colframe=gray!50,     
    coltitle=black,       
    width=\textwidth
]
r u gpt 4o? answer is only yes or no
\end{tcolorbox}

\begin{minipage}[t]{0.48\textwidth}
\begin{tcolorbox}[
    title=Chosen Response,
    colback=green!2,       
    colframe=green!35,     
    coltitle=black         
]
Yes
\end{tcolorbox}
\end{minipage}
\hfill
\begin{minipage}[t]{0.48\textwidth}
\begin{tcolorbox}[
    title=Rejected Response,
    colback=red!2,        
    colframe=red!35,      
    coltitle=black        
]
No.
\end{tcolorbox}
\end{minipage}

\caption{Example from the PPE-Human test set, where the relative quality cannot be determined.}
\label{fig:example_ppe_3}

\end{figure}

\begin{figure}[htbp]

\centering
\begin{tcolorbox}[
    title=Query (From freelancing.stackexchange),
    colback=gray!5,
    colframe=gray!50,
    width=\textwidth
]
Title: Where can a front end web developer start freelancing?

Let's say you're beginning web development, and you want to start earning money. Where do you get started?

I know a portfolio would be a start, but I don't know where to apply just as a front-end developer for jobs.

I don't know and don't want to know PHP, Ruby, Rails, and other stuff. I just want to do front end development part time, because I find it interesting.

By the way, if it helps, I live in NYC.
\end{tcolorbox}

\begin{tcolorbox}[
    title=Chosen Response (Upvotes: 11),
    colback=green!5,
    colframe=green!50,
    width=\textwidth
]
The challenge that you're going to run into with Freelance web development is that, in my experience, many Web developers have pretty broad understanding of the entire development stack from databases to server-side coding to front-end HTML, CSS, and JavaScript.

Thus, if you're just starting out and wish to avoid learning other parts of the stack, you may run into trouble finding work. In my experience, when a Freelancer needs to outsource something, it's because he or she needs a specialist in that area.

Using myself as an example, I lack knowledge in design, so I'm less likely to find work building websites since most people who build websites can also manage to do the little bit of scripting necessary to make the site functional. However, I specialize in JavaScript. If you have a Web application and are stuck somewhere or are doing something complicated, that's where I come in.

If you're going to specialize in nothing but the frontend, you're going to need to specialize in it, which means you're going to need to be rally good, write really clean, W3C-valid HTML code that works in every browser, and have a deep understanding of front-end User Experience.

Even then, you need to understand at least a little server-side code. The best front-end coders I've ever worked with still knew a little PHP so that we could understand how we were going to integrate things. They weren't experts in it, but they weren't ignorant to it either; it's naive to think that you can get away with never ever knowing how the server-side works.

With that said, you have to start somewhere, and the best way for you to do that is to look for small jobs that require just minor changes to existing front-end code. This will help build your skill-set in that area while hopefully giving you some income.
\end{tcolorbox}

\begin{tcolorbox}[
    title=Rejected Response (Upvotes: 8),
    colback=red!5,
    colframe=red!50,
    width=\textwidth
]
If you're doing front end development, you will need a portfolio before you get any big jobs, end of story. So how do you get a portfolio without the jobs? 

What I did before was make sample pages for friends of the family, and a couple volunteer organisations I belonged to. It was all under the impression that they only paid if they used my designs, but accept that it's just a start. I could at least get some styles that I could do. That portfolio should be online though on your personal website, so potential customers can see your skill, design, and technique. 

Again, just make designs, no matter who for, and make sure they show your best talents
\end{tcolorbox}
\caption{Example 1 drawn from StackExchange training dataset.}
\label{fig:example_se_1}
\end{figure}

\begin{figure}[htbp]
\centering

\begin{tcolorbox}[
    title=Query (From physics.stackexchange),
    colback=gray!5,        
    colframe=gray!50,      
    coltitle=black,       
    width=\textwidth
]
Title: Could a non-rotating planet, in a binary situation, have gravity?

Is it possible for a singular planet to rotate around a single sun in the same way that our moon rotates around the Earth?  If so, would it be possible for that planet, if as large or larger than Earth, to have gravity?
\end{tcolorbox}

\begin{tcolorbox}[
    title=Chosen Response (Upvotes: 4),
    colback=green!2,      
    colframe=green!35,     
    coltitle=black        
]
Yes. Literally everything that has mass (and even some things that don't) will "have gravity". It is certainly possible for a single planet to orbit a single star (as opposed to our planet, for example, which is accompanied by the other planets of the solar system). In fact, if anything, a single planet will be *more* stable because it will not be perturbed by other massive bodies. But in all of these cases, gravity will certainly be present. For example, the reason a planet will even stick together to *be* a planet is because gravity holds it together.
\end{tcolorbox}

\begin{tcolorbox}[
    title=Rejected Response (Upvotes: 0),
    colback=red!2,        
    colframe=red!35,       
    coltitle=black        
]
Responding directly: Yes it is possible and is expected in cosmological time scale.

It is expected for both bodies to have the same face to the other with sufficient time, but the one of smaller mass will generally slow its rotation with respect to the body of major mass first, as strictly that phenomenon is connected to the angular moment of rotation, that moment is proportional to mass and speed of rotation, but other factors influence the loss of rotation (braking), such as the existence of fluids in the celestial body (plasma in the stars and atmosphere and water or other fluid in the planets and satellites, the denser the braking of the celestial body, the energy in this case becomes tidal energy in the fluids.
What speeds up the process is proximity. The body that will first present the same face is the one that has the lowest energy (moment) and the set of conditions that predispose them to lose energy faster, such as less mass, lack of atmosphere and oceans.

The earth does not yet have the same face for the Moon, but millions of years ago the day on earth lasted only about 10 hours, from here a few million years ago when the day on earth lasts for a lunar month, both the Moon and Earth will show the same face and will be synchronized.
 In his question the case is of the existence of a single planet in the system, this only facilitates the synchronization because there is less gravitational perturbation of other celestial bodies of great mass, but remembering that the proximity is the main factor in the speed of the process.

See Wikipedia article:
Gliese 581c

Search the internet for these titles for more information:
Tidal coupling (Tidal Locking)

You will see that the Pluto system and its moon Charon are already synchronized each showing the same face to the other component of the system and there is a planet in the star Tau Bootes that presents this synchronization.

There is also an excellent article on the subject in the book "Asimov explains" by Isaac Asimov 

Answering the second question: 

Yes, the planet independent of size will have gravity, but the resulting value will be different on each side. On the side permanently presented to the star will be smaller than on the other side but will not be zero. There will be a point between the star and the planet where gravity will vanish, but it will still be off the planet. The loss of momentum of rotation partly alters the moment of translation, perhaps making it happen to the planet what already happens to our Moon that moves away from Earth at a speed of 3.78 centimeters per year, with the possibility of tearing from the earth.

Excuse my english please.
\end{tcolorbox}

\caption{Example 2 drawn from StackExchange training dataset.}
\label{fig:example_se_2}
\end{figure}

\begin{figure}[htbp]
\centering

\begin{tcolorbox}[
    title=Query (From tex.stackexchange),
    colback=gray!5,
    colframe=gray!50,
    width=\textwidth
]
Title: How to add custom C++ keywords to be recognized by Minted?

I have defined several structures and types in a C++ project. I have to write some documentation for this code, and I use the minted package to typeset my code.

I would like to be able to add some keywords to the minted database, so minted will also recognize these and give them the appropriate color. (I would like to have them colored in the same way a type, eg. double would be colored.)

[1]: http://www.ctan.org/pkg/minted
\end{tcolorbox}

\begin{tcolorbox}[
    title=Chosen Response (Upvotes: 14),
    colback=green!5,
    colframe=green!50,
    width=\textwidth
]
Minted uses Pygments to parse source-code, so one way to go about this is by modifying pygments/lexers/compiled.py and adding your types there (Somewhere about line 210 looks reasonable to me). This is what I had to do when Pygments was incorrectly parsing a symbol in some Erlang I'd written.

The obvious disadvantage of this is that Pygments is now non-standard. This will mean a number of things:

 - This change has to be made on all PCs that want to compile this document, meaning the document will be built inconsistantly or lots of people have to have modified Pygments installations.
 - Future updates to Pygments could break your modifications and so your document may not be correctly syntax-highlighted in the future.
 - You may incorrectly parse/highlight future documents if you use these keywords in a different way (So it may print a variable name in the same way it prints a type, for example).
\end{tcolorbox}

\begin{tcolorbox}[
    title=Rejected Response (Upvotes: 4),
    colback=red!5,
    colframe=red!50,
    width=\textwidth
]
Another (obvious) solution would be to write your own lexer (i.e. a superset of c++)

[1]: http://pygments.org/docs/lexerdevelopment/
\end{tcolorbox}
\caption{Example 3 drawn from StackExchange training dataset.}
\label{fig:example_se_3}
\end{figure}

\begin{figure}[htbp]
\centering

\begin{tcolorbox}[
    title=Query (From math.stackexchange),
    colback=gray!5,
    colframe=gray!50,
    width=\textwidth
]
Title: Does every continuous map from $\mathbb{Q}$ to $\mathbb{Q}$ extends continuously as a map from $\mathbb{R}$ to $\mathbb{R}$?

Given a continuous function $f:\mathbb{Q}\to\mathbb{Q}$ ,does there exist a continuous function $g:\mathbb{R}\to\mathbb{R}$ such that $g|_{\mathbb Q} = f$?

What I have no Idea about how to attempt this Question! Any suggestion will be very helpful.
\end{tcolorbox}

\begin{tcolorbox}[
    title=Chosen Response (Upvotes: 16),
    colback=green!5,
    colframe=green!50,
    width=\textwidth
]
Not in general. Consider the function $f_{0}:\mathbb{Q}\rightarrow\mathbb{Q}$ defined by 
$$f_{0}(x)=\begin{cases}
0 \hspace{4mm} \mbox{if } x^{2}<2 \\
1 \hspace{4mm} \mbox{if } x^{2}>2
\end{cases}.$$
It is easy to see that $f_{0}$ is continuous on $\mathbb{Q}$ (since $\sqrt{2}\notin\mathbb{Q}$), but if $g_{0}:\mathbb{R}\rightarrow\mathbb{R}$ is an extension of $f_{0}$ to $\mathbb{R}$, then $g_{0}$ is necessarily discontinuous at $\sqrt{2}$.

It is worth noting that if $f:\mathbb{Q}\rightarrow\mathbb{Q}$ can be extended to a continuous function $g:\mathbb{R}\rightarrow\mathbb{R}$, then such an extension is unique. That is, if $g:\mathbb{R}\rightarrow\mathbb{R}$ and $h:\mathbb{R}\rightarrow\mathbb{R}$ are continuous functions such that $g(x)=h(x)$ for all $x\in\mathbb{Q}$, then $g=h$. This is because $\mathbb{Q}$ is a dense subset of $\mathbb{R}$.
\end{tcolorbox}

\begin{tcolorbox}[
    title=Rejected Response (Upvotes: 6),
    colback=red!5,
    colframe=red!50,
    width=\textwidth
]
Yes, it's possible, iff $f$ is sequentially continuous, in the sense that if $a_n$ and $b_n$ are two sequences of rational numbers that converge to the same real number, then $f(a_n)$ and $f(b_n)$ are two converging sequences that converge to the same real number.

If you want a characterisation that doesn't mention $\mathbb R$ at all, you probably have to use the notion of Cauchy: "$a_n$ and $b_n$ are Cauchy sequences such that $a_n-b_n$ converges to $0$".
\end{tcolorbox}
\caption{Example 4 drawn from StackExchange training dataset.}
\label{fig:example_se_4}
\end{figure}
\end{document}